\documentclass[]{antgroup}
\PassOptionsToPackage{numbers, compress}{natbib}
\usepackage{antgroup}

\usepackage{graphicx}
\usepackage{tikz}
\usepackage{todonotes}
\usepackage{multirow}
\usepackage{amsmath}
\usepackage{cleveref}
\usepackage{subcaption}
\usepackage{multicol}
\usepackage{amssymb}
\usepackage{booktabs}
\usepackage{array}
\usepackage{bm}
\usepackage{enumitem}
\usepackage{algorithm}
\usepackage{algpseudocode}
\usepackage{tabularx}
\usepackage{bbm}
\usepackage{makecell}
\usepackage{threeparttable}
\definecolor{title_blue}{HTML}{1677FF} 
\definecolor{cite_blue}{HTML}{044dc1}
\usepackage{colortbl}

\usepackage[normalem]{ulem}
\useunder{\uline}{\ul}{}

\usepackage{multirow}

\usepackage{amsmath,amsfonts,bm}

\def\eqref#1{equation~\ref{#1}}

\def\1{\bm{1}}

\DeclareMathAlphabet{\mathsfit}{\encodingdefault}{\sfdefault}{m}{sl}
\SetMathAlphabet{\mathsfit}{bold}{\encodingdefault}{\sfdefault}{bx}{n}

\newcommand*\justify{%
  \fontdimen2\font=0.4em
  \fontdimen3\font=0.2em
  \fontdimen4\font=0.1em
  \fontdimen7\font=0.1em
  \hyphenchar\font=`\-
}

\renewcommand{\texttt}[1]{%
  \begingroup
  \ttfamily
  \begingroup\lccode`~=`/\lowercase{\endgroup\def~}{/\discretionary{}{}{}}%
  \begingroup\lccode`~=`[\lowercase{\endgroup\def~}{[\discretionary{}{}{}}%
  \begingroup\lccode`~=`.\lowercase{\endgroup\def~}{.\discretionary{}{}{}}%
  \catcode`/=\active\catcode`[=\active\catcode`.=\active
  \justify\scantokens{#1\noexpand}%
  \endgroup
}

\usepackage{amsmath}
\usepackage{amssymb}
\usepackage{amsfonts}                               
\usepackage{amsthm}
\usepackage[mathcal]{eucal}
\usepackage{mathrsfs}
\usepackage{bm}                                     
\usepackage{blkarray}                               
\usepackage{nicefrac}                               

\usepackage{wrapfig}
\usepackage{graphicx}                               
\usepackage{caption}
\usepackage{cleveref}

\usepackage{tikz}                                          
\usepackage{circuitikz}
\usetikzlibrary{patterns,snakes}
\usetikzlibrary{positioning,calc,fit,decorations.pathmorphing,shapes.geometric, shapes.gates.logic.US, calc}
\usetikzlibrary{arrows,arrows.meta,decorations.markings,shapes,shapes.arrows}
\usetikzlibrary{decorations,decorations.pathreplacing}
\usetikzlibrary{backgrounds}
\usepackage{filecontents}                           
\usepackage{pgfplots}
\usepackage{pgfplotstable}
\usepgfplotslibrary{groupplots}
\usepackage{scalefnt}
\pgfplotsset{compat=newest}

\usepackage{xcolor}
\definecolor{firstcolor}{HTML}{C3423F}
\definecolor{secondcolor}{HTML}{2A4B8C}
\definecolor{myblue}{HTML}{D9E6FF}

\newcommand{\github}{\raisebox{-1.5pt}{\includegraphics[height=1.05em]{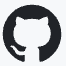}}}
\newcommand{\hf}{\raisebox{-1.5pt}{\includegraphics[height=1.05em]{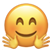}}}

\usepackage{tocloft}
\title{LLaDA2.0-Uni: Unifying Multimodal Understanding and Generation with Diffusion Large Language Model}

\author{
\centering
Tiwei Bie,
Haoxing Chen,
Tieyuan Chen,
Zhenglin Cheng,
Long Cui,
Kai Gan,
Zhicheng Huang\\
Zhenzhong Lan$^{\dag}$,
Haoquan Li,
Jianguo Li$^{\dag}$,
Tao Lin$^{\dag}$,
Qi Qin,
Hongjun Wang,
Xiaomei Wang\\
Haoyuan Wu,
Yi Xin, 
Junbo Zhao
}

\footnotetext{Authors are listed in alphabetical order based on last name. $\dag$ indicates tech-leaders.}

\affiliation{AGI Research Center, Inclusion AI}

\begin{document}
\maketitle

\begin{abstract}
We present LLaDA2.0-Uni, a unified discrete diffusion large language model (dLLM) that supports multimodal understanding and generation within a natively integrated framework. 
Its architecture combines a fully semantic discrete tokenizer, a MoE-based dLLM backbone, and a diffusion decoder. 
By discretizing continuous visual inputs via SigLIP-VQ, the model enables block-level masked diffusion for both text and vision inputs within the backbone, while the decoder reconstructs visual tokens into high-fidelity images. 
Inference efficiency is enhanced beyond parallel decoding through prefix-aware optimizations in the backbone and few-step distillation in the decoder. 
Supported by carefully curated large-scale data and a tailored multi-stage training pipeline, LLaDA2.0-Uni matches specialized VLMs in multimodal understanding while delivering strong performance in image generation and editing. 
Its native support for interleaved generation and reasoning establishes a promising and scalable paradigm for next-generation unified foundation models.

\begin{center}
    \renewcommand{\arraystretch}{1.3}
    \begin{tabular}{rll}
        \github{} & \textbf{GitHub} & \url{https://github.com/inclusionAI/LLaDA2.0-Uni} \\
        \hf{} & \textbf{HuggingFace Model} & \url{https://huggingface.co/inclusionAI/LLaDA2.0-Uni} 
    \end{tabular}
\end{center}

\end{abstract}

\begin{figure}[htb!]
    \centering
    \vspace{-0.2in}
    \includegraphics[width=0.95\linewidth]{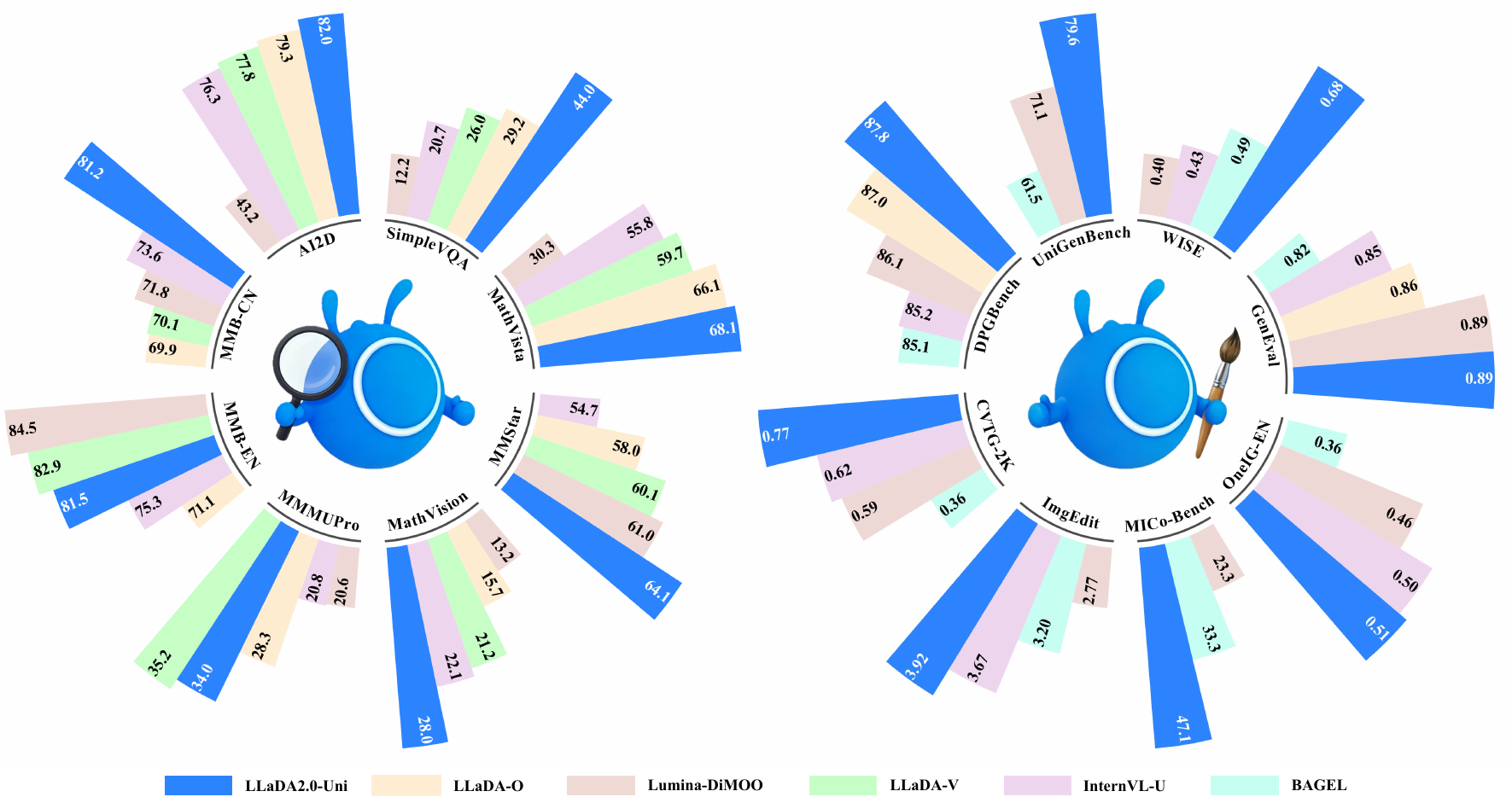} 
    \vspace{-0.05in}
    \caption{Benchmark Performance of LLaDA2.0-Uni.}
    \label{fig:main}
\end{figure}

\newpage
\begin{figure}[H]
  \vspace{-0.88in}
  \hspace{-0.05in}
  \includegraphics[width=1.0\textwidth]{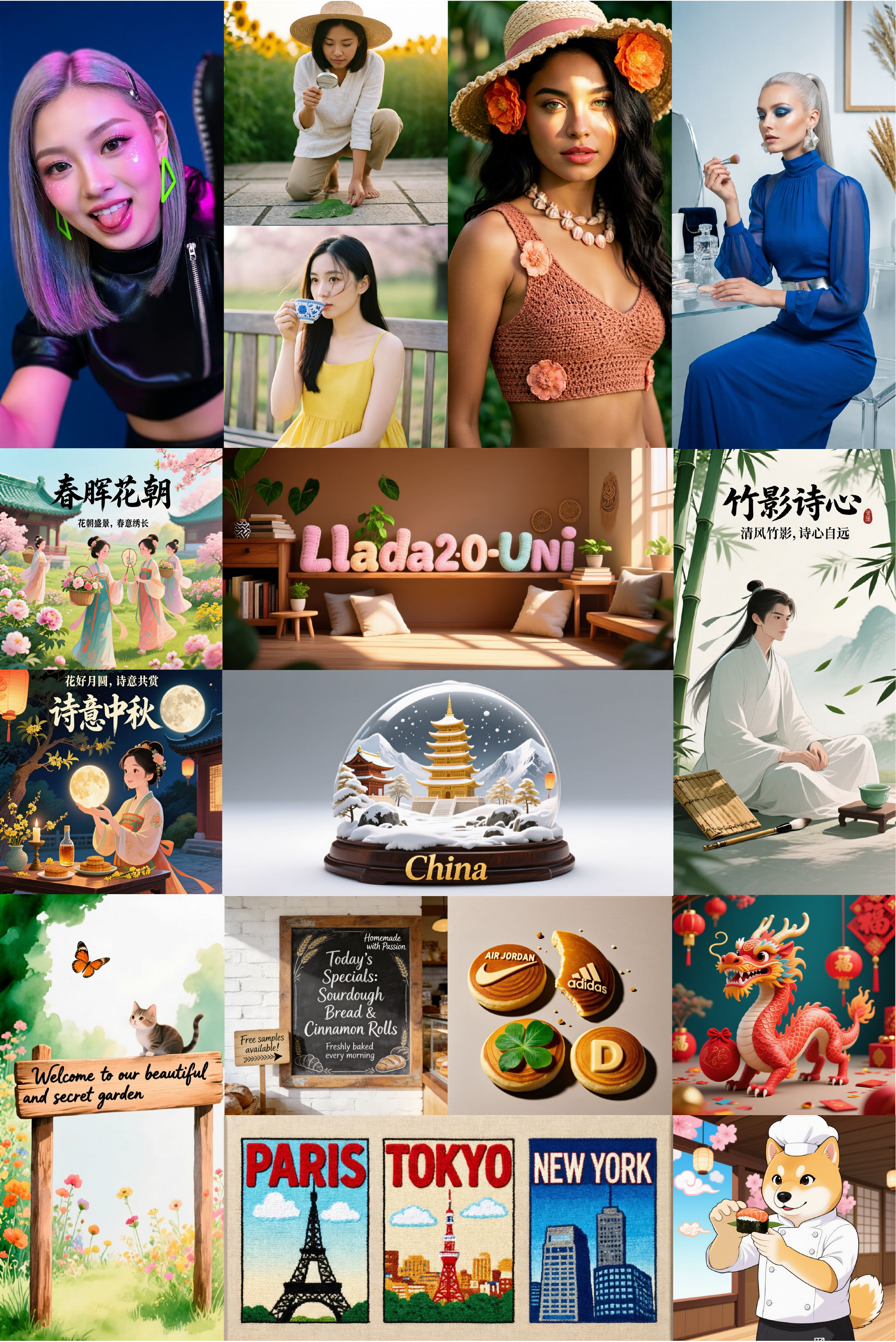}
  \caption{Showcases of LLaDA2.0-Uni in High-Fidelity Image Generation.}
  \label{fig:showcase_realistic}
\end{figure}
\newpage

\begin{figure}[H]
  \vspace{-0.88in}
  \hspace{-0.05in}
  \includegraphics[width=1.0\textwidth]{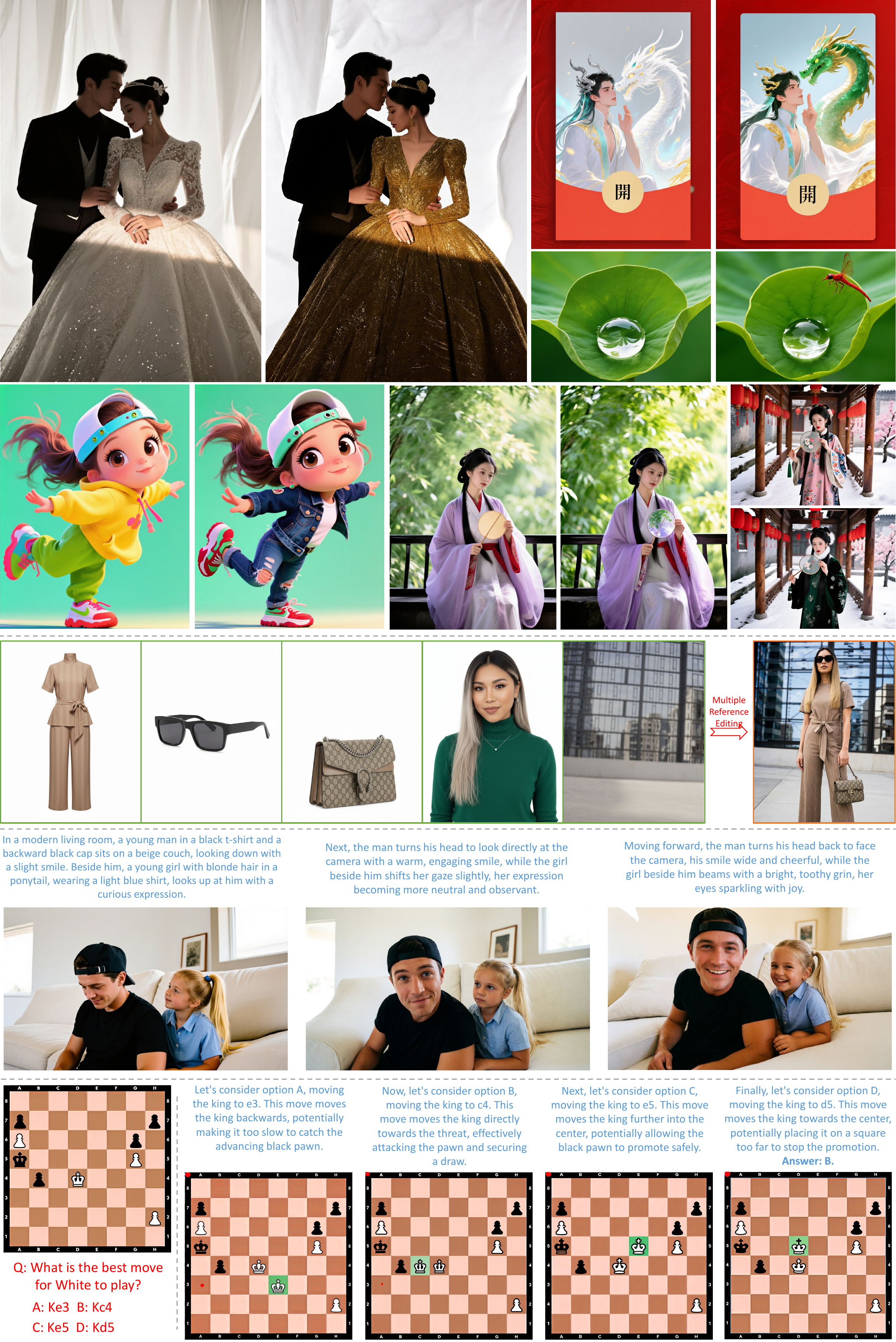}
  \vspace{-0.27in}
  \caption{Showcases of LLaDA2.0-Uni in Single/Multi-Reference Editing, Interleaved Generation and Reasoning. Note: For editing tasks, the original image is positioned on the left or top.}
  \label{fig:showcase_editing_interleaved}
\end{figure}

\clearpage


\newpage
\section{Introduction}
\label{sec:intro}
Large language models (LLMs) have evolved beyond text to handle a wide variety of multimodal tasks~\citep{cui2025emu3,ming_flash,wu2025qwen,gu2025ui,liu2026lumina}.
Understanding and generation represent the two primary categories of multimodal tasks. 
By modeling both visual understanding and generation as token sequence prediction, LLMs have achieved highly competitive results in both areas. 
Traditionally, these tasks are handled by separate specialized models, such as Qwen-VL~\citep{bai2025qwen2,team2025qwen3VL} or InternVL~\citep{Internvl,internvl2,internvl35} for understanding, and Flux~\citep{flux} or Z-Image~\citep{cai2025z} for generation. 
However, a unified model that handles both within a single framework offers several key benefits: it promotes mutual enhancement between understanding and generation, improves deployment efficiency, and unlocks advanced capabilities like interleaved generation and reasoning, ultimately bringing us closer to artificial general intelligence (AGI).

Current unified multimodal models predominantly build upon autoregressive (AR) architectures.
Janus~\citep{Janus} and Lumina-mGPT~\citep{liu2026lumina} tokenizes images into discrete sequences and unifies both modalities under next-token prediction, while OmniGen2~\citep{wu2025omnigen2}, Hunyuan Image 3.0~\citep{cao2025hunyuanimage}, and BAGEL~\citep{bagel} adopt a hybrid paradigm combining text autoregression with image diffusion.
While these AR-based approaches have shown promise, masked diffusion models~\citep{lou2023discrete,sahoo2024simple,xin2025resurrect} offer an alternative paradigm with inherent advantages in parallel decoding and bidirectional context modeling.
A unified masked diffusion framework further simplifies training through a single objective, avoiding the delicate balance between AR and diffusion losses.
However, existing unified masked diffusion models, such as MMaDA~\citep{Mmada} and Lumina-DiMOO~\citep{dimoo}, still lag behind state-of-the-art AR-based unified architectures in both task coverage and benchmark performance.
This gap fundamentally stems from their architecture and modeling designs: 1) their reconstructive VQ tokenizers lack semantic information, causing poor understanding performance; 2) excessive image compression by VQ tokenizers compromises generation quality; 3) their fully bidirectional modeling has been shown to be unreliable for text.
Furthermore, they commonly assume fixed output lengths for understanding tasks, limiting their applicability in open-ended scenarios.

To overcome these limitations, we propose LLaDA2.0-Uni, a unified dLLM-based Mixture-of-Experts (MoE) model for seamless multimodal understanding and generation.
At its core, LLaDA2.0-Uni utilizes LLaDA2.0~\citep{LLaDA2} (a 16B dLLM MoE architecture) as its backbone.
A key architectural innovation is the introduction of the SigLIP-VQ tokenizer, which converts continuous visual inputs into fully discrete semantic tokens.
Unlike previous reconstruction-based tokenizers that struggle with multimodal understanding, this purely semantic representation preserves crucial details and effectively supports complex visual reasoning.
Consequently, this design maintains the unified discrete modeling format, allowing both text and images to be optimized under a shared block-level masked diffusion objective.
For image generation, LLaDA2.0-Uni employs a dedicated Diffusion Decoder to process the discrete tokens generated by the dLLM backbone.
Optimized through distillation, this decoder synthesizes high-fidelity images in just 8 inference steps, achieving an excellent balance between speed and quality.

LLaDA2.0-Uni achieves top-tier performance across both understanding and generation benchmarks, as shown in Figure~\ref{fig:main}.
In multimodal understanding, LLaDA2.0-Uni demonstrates competitive visual question answering and document reasoning capabilities compared with specialized VLMs such as Qwen2.5-VL~\citep{bai2025qwen2}
Regarding image generation, LLaDA2.0-Uni produces high-quality images and enables highly flexible image editing.
Beyond these general tasks, the unified discrete representation empowers LLaDA2.0-Uni to support interleaved generation and reasoning.
This flexibility establishes LLaDA2.0-Uni as a powerful and efficient paradigm for the next generation of unified foundation models.

The key contributions of LLaDA2.0-Uni can be summarized as follows:

\begin{itemize}[leftmargin=16pt]
    \item \textbf{Novel Unified Architecture.} LLaDA2.0-Uni integrates a fully semantic tokenizer, a 16B MoE dLLM backbone, and a diffusion decoder. This architecture unifies text and image modeling through a shared block-wise mask prediction objective.
    
    \item \textbf{Interleaved Generation and Reasoning.} Beyond its strong performance in both understanding and generation, LLaDA2.0-Uni inherently supports interleaved generation and reasoning, marking a significant step toward exploring how generation and understanding can reinforce each other.
    
    \item \textbf{Efficient Inference.} Building on the advantages of parallel decoding, LLaDA2.0-Uni further accelerates inference by optimizing the decoding process in the dLLM backbone and applying few-step distillation to the decoder, achieving an effective balance between speed and performance.

    \item \textbf{Strong Benchmark Performance.} LLaDA2.0-Uni achieves strong performance across visual understanding, generation, and editing benchmarks, performing on par with state-of-the-art unified models.
\end{itemize}

\begin{figure}[tb]
    \centering
    \includegraphics[width=1.0\linewidth]{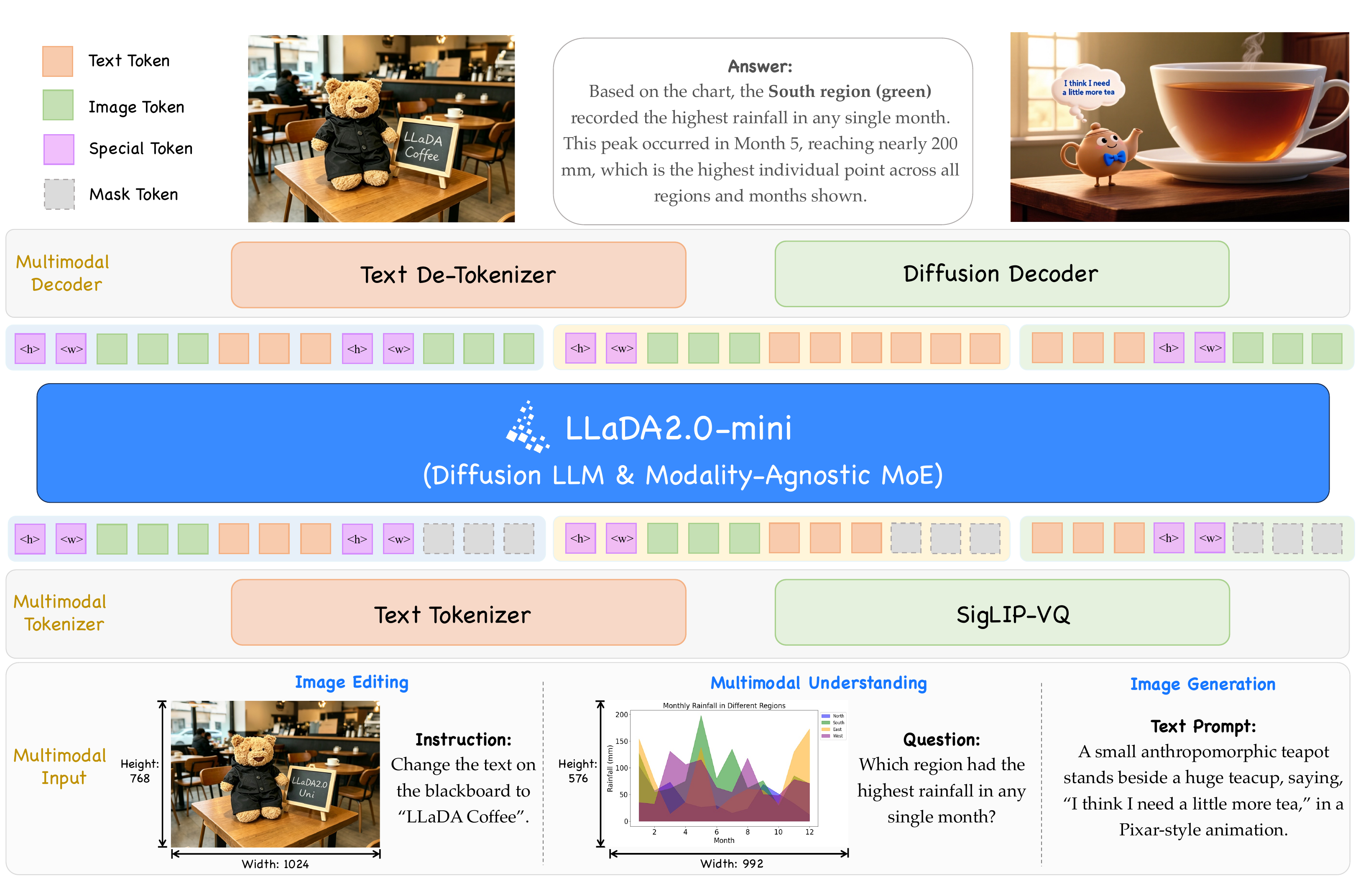}
    \vspace{-0.25in}
    \caption{Architecture Overview of LLaDA2.0-Uni.
    The framework integrates a SigLIP-VQ tokenizer and a language model decoder to process multimodal inputs, including text, image, and video.
    }
    \label{fig:unillada}
\end{figure}

\section{Model Design}
\subsection{Motivation and Design Principles}
We aim to extend the dLLM architecture into a unified model for multimodal understanding and generation, leveraging its advantages in parallel decoding and bidirectional context modeling.
Prior approaches to this goal exhibit notable limitations.
MMaDA~\citep{Mmada} and Lumina-DiMOO~\citep{dimoo} rely on reconstructed visual tokens from VQ-VAE, which degrades understanding performance and yields sub-optimal visual quality.
LLaDA-o~\citep{you2026lladao} and BAGEL~\citep{bagel} adopt decoupled visual modules (ViT for understanding, VAE for generation), introducing a modeling gap and divergent optimization objectives within the same model.
To overcome these limitations, we design LLaDA2.0-Uni around a key principle: \emph{use fully discrete semantic tokens for both understanding and generation}.
This unified representation eliminates the need for heterogeneous encoders and enables end-to-end training under a single mask prediction objective, as illustrated in Figure~\ref{fig:unillada}.

\subsection{Architecture}
LLaDA2.0-Uni consists of three core components: (1) a SigLIP-VQ tokenizer that converts images into discrete semantic tokens, (2) a 16B MoE diffusion language model that processes both text and visual tokens under a unified mask prediction objective, and (3) a diffusion decoder that reconstructs semantic tokens into high-fidelity images.
This design enables end-to-end training and inference for both understanding and generation tasks within a single coherent framework.

\subsubsection{Semantic Discrete Tokenizer}
The tokenizer adopts a SigLIP-VQ architecture building upon X-Omni~\citep{xomni} to convert continuous images into discrete tokens. 
Unlike standard VQ-VAEs~\citep{esser2021taming,wang2024emu3} that rely on pixel-level reconstruction, SigLIP-VQ is trained directly on understanding tasks, thereby preserving rich semantic information.
Consequently, it demonstrates a clear advantage over reconstruction-based VQ-VAEs in multimodal understanding tasks.
Specifically, the tokenizer utilizes a pre-trained SigLIP2-g ViT~\citep{tschannen2025siglip} as the visual feature extractor and supports dynamic resolution processing.
Following the ViT encoder, a vector quantizer aligns the visual representations with a pre-trained large language model, featuring a codebook with a vocabulary size of 16,384 and a dimensionality of 2,048.
While SigLIP-VQ excels in semantic extraction, it lacks a native mechanism to reconstruct images from these discrete tokens.
We address this by designing a custom diffusion decoder, detailed in Section~\ref{sec:diffusion_decoder}.

\subsubsection{Diffusion Large Language Model}
\paragraph{MoE Backbone for Multi-Modal Capacity.}
A modality-agnostic Mixture-of-Experts (MoE) architecture enables language backbones to serve as universal multi-task learners, dynamically allocating capacity across modalities without the need for modality-specific designs.
We adopt LLaDA-2.0-mini~\citep{LLaDA2} as our dLLM backbone, an MoE architecture with 16B total parameters.
To integrate visual information, we expand the original dLLM vocabulary by appending tokens from the SigLIP-VQ codebook, along with a set of custom special tokens for image generation and understanding.
In the input embedding layer, we retain the pre-trained language embeddings while randomly initializing the new visual token embeddings.
Similarly, the final prediction head is expanded to accommodate the enlarged vocabulary, with the language-specific portion initialized from pre-trained weights to preserve linguistic proficiency.

\paragraph{Block-wise Attention for Training Stability.}
For dLLMs, full bidirectional attention is theoretically ideal for parallel sampling.
However, prior studies~\citep{LLaDA,Mmada,dimoo} show that unconstrained full attention often degrades performance.
We adopt a block-wise attention scheme~\citep{arriola2025block} to balance quality and efficiency.
This design is particularly important for SigLIP-VQ tokens: since they are semantically aligned with Qwen2.5, they inherit an autoregressive bias that would be disrupted by pure full-attention.
By constraining attention within predefined blocks and selectively enabling it across blocks, we maintain parallel decoding speed while achieving strong performance in both language and visual tasks.
\looseness=-1

\paragraph{Positional Embedding \& Arbitrary Resolution.} 
Rotary Position Embedding (RoPE)~\citep{su2021roformer} is a standard choice in LLMs due to its flexibility and scalability. 
While many recent unified models adopt 2D RoPE for images~\citep{cao2025hunyuanimage,bagel}, we keep the original 1D RoPE structure for simplicity.
To represent 2D spatial information, we add special \texttt{<height>} and \texttt{<width>} tokens (e.g., \texttt{<imgsize\_512>}) before the flattened 1D visual sequence.
Previous studies~\citep{liu2026lumina,xin2025lumina,xomni} confirm that this simple approach is highly effective.
These size tokens also enable the model to handle arbitrary image resolutions without architectural changes.

\subsubsection{Diffusion Decoder}
\label{sec:diffusion_decoder}
Semantic VQ requires a specialized decoder to map features from the semantic space back to the image space, unlike traditional reconstruction-based VQ that can directly use a pixel decoder.
We introduce a diffusion model built upon Z-Image-Base~\citep{cai2025z}, a 6B pre-trained text-to-image model.
Once the dLLM generates image tokens, they serve as the conditioning signal, replacing conventional text prompts.
This differs from existing methods like NextFlow~\citep{zhang2026nextflow} and X-Omni~\citep{xomni}, which redundantly combine text prompts with visual tokens.
Beyond basic decoding, our diffusion model performs $2\times$ super-resolution, using upsampled semantic tokens as the sole conditioning input.
To address the computational cost of 50-step sampling with CFG, we employ model distillation to achieve 8-step CFG-free inference (Section~\ref{sec:decoder_training}).
Together, the SigLIP-VQ tokenizer, dLLM backbone, and diffusion decoder form a unified pipeline where understanding and generation share the same discrete token representation.

\subsection{Training-free Inference Acceleration}
\label{sec:sprint}

Block-wise discrete diffusion language models require $B \times T$ forward passes to generate $B$ blocks with $T$ denoising steps each. Uniform KV cache eviction and fixed-schedule step reduction degrade quality in multimodal settings due to heterogeneous per-token difficulty and differing information density across modalities. We propose \textbf{SPRINT} (\textbf{S}parse \textbf{P}refix \textbf{R}etention with \textbf{I}nference-time \textbf{N}on-uniform \textbf{T}oken Unmasking), a training-free framework that reduces cost along two orthogonal axes. \emph{Sparse Prefix Retention} prunes the prefix KV cache in a modality-aware manner to lower per-step cost. \emph{Non-uniform Token Unmasking} replaces the fixed denoising schedule with confidence-adaptive unmasking to reduce the step count. Together the two components achieve up to $1.6\times$ speedup with negligible quality loss (Section~\ref{sec:sprint_ablation}).

\noindent
\textbf{Sparse Prefix Retention.}\label{sec:sprint_spr}
Each denoising step attends to the full prefix, whose quadratic attention cost dominates as the generated sequence lengthens.
SPRINT constructs a pruned prefix KV cache once per block, so that all subsequent steps attend to a much shorter effective sequence.

The first step of each block performs a full forward pass to obtain logits and a complete KV cache.
Each prefix position $i$ is then scored by a composite importance measure that blends the \emph{key-norm importance} $\bar{I}_i$, reflecting how strongly a position influences the attention distribution, with the \emph{token confidence} $c_i$, capturing the model's prediction certainty at that position:
\begin{equation}\label{eq:sprint_importance}
    s_i \;=\; \alpha \cdot \bar{I}_i \;+\; (1 - \alpha) \cdot c_i \,,
\end{equation}
where $\bar{I}_i = {\|\mathbf{k}_i\|_2}\big/\bigl({\frac{1}{L}\sum_{j=1}^{L}\|\mathbf{k}_j\|_2}\bigr)$ is the mean-normalized key norm, $c_i = \max_{v}\, p_\theta(v \mid \mathbf{x}_t)$ is the top-1 softmax confidence, and $\alpha = 0.5$.

The pruning is \emph{modality-aware}: we maintain separate keep ratios $r_{\text{text}}$ and $r_{\text{img}}$ rather than a single uniform ratio, because image tokens exhibit high spatial redundancy and tolerate aggressive pruning, whereas text tokens carrying instructions or reasoning chains do not.
Within each modality, the top-$\lfloor r \cdot n \rfloor$ positions by $s_i$ are retained and the rest are evicted.
We explore two settings: \emph{selective image pruning} with $r_{\text{text}} = 1.0, r_{\text{img}} = 0.8$, and \emph{full prefix retention} with $r_{\text{text}} = r_{\text{img}} = 1.0$.
The global keep ratio is $r = 0.5$ in both cases.

\noindent
\textbf{Non-uniform Token Unmasking.}\label{sec:sprint_ntu}
The standard denoising schedule unmasks a fixed $\lceil m / T \rceil$ tokens per regardless of prediction certainty, wasting computation on confident predictions and under-allocating it to uncertain ones.
SPRINT replaces this schedule with a confidence-adaptive strategy. For all $m$ masked positions the model computes per-position confidence $c_n = p_\theta(\hat{x}_0^n \mid \mathbf{x}_t)$ and accepts every position whose confidence exceeds a threshold $\tau$ in a single step.
\begin{equation}\label{eq:adaptive_fast}
    \mathcal{A} \;=\; \bigl\{\,n \in [m] : c_n > \tau \,\bigr\} \,.
\end{equation}
A minimum of $\lceil m / (T - t) \rceil$ acceptances is enforced at each step to guarantee termination. We examine $\tau \in \{0.93, 0.95\}$.

\section{Data Preparation}
\subsection{Multimodal Understanding}
\paragraph{Pretrain Data Source \& Processing.} 
In the pre-training stage, the model learns to perceive images through text supervision.
We collect extensive image-captioning data from open-source datasets~\citep{LLaVA-OneVision-1.5,Penguin-VL}, supplemented by specialized categories:
\begin{itemize}[leftmargin=16pt]
    \item \textbf{OCR Data.} We develop a coarse-to-fine pipeline to produce millions of samples. By combining PaddleOCR~\citep{Paddleocr} pseudo-labels with refinements from Qwen3-VL, we achieve high-quality document understanding data without manual annotation.

    \item \textbf{Grounding \& Counting Data.} Using Objects365~\citep{Objects365} and RefCOCO~\citep{yu2016modeling,openimages,kazemzadeh2014referitgame}, we refine spatial data through detection confidence filtering and Qwen3-VL-235B-A22B \citep{team2025qwen3VL} verification. 
    Coordinates are normalized to [0, 1000] for stability, and counting data is automatically derived from verified bounding boxes.

    \item \textbf{World Knowledge \& Reasoning.} We curate data across three domains: general world knowledge, logical reasoning, and mathematics.

    \item \textbf{Text Data.} High-quality text-only data is sourced from Ling2.0~\citep{Ling2} and LLaDA2.0~\citep{LLaDA2}, covering general knowledge, code, and mathematics.
\end{itemize}

\paragraph{SFT Data Source \& Processing.} 
Our SFT dataset contains approximately 60 million samples with a 1:5 ratio of text-only to multimodal data.
This collection covers single/multi-turn dialogues and single/multi-image scenarios across various tasks, including General VQA, Chart/Table QA, mathematical reasoning, etc.
We implement a two-stage filtering pipeline for quality control:
(1) \textbf{Query Filtering:} Qwen3-VL audits the input space, pruning vague or low-information instructions while rewriting ambiguous queries to enhance semantic clarity.
(2) \textbf{Response Filtering:} Rule-based heuristics rectify structural artifacts, and GPT-OSS~\citep{agarwal2025gpt} filters semantic biases while ensuring alignment with ground-truth references.

\subsection{Image Generation}
\paragraph{Data Source.}
We collect over 200 million web images with their original text descriptions.
To improve performance on challenging generation tasks, we specifically increased the proportion of images featuring human body and rendered text.
Since image generation requires higher visual quality than understanding, all data undergoes rigorous filtering.

\paragraph{Filtering Pipeline.}
We apply a three-stage cleaning process:
(1) \textbf{Metadata filtering} removes low-resolution images (less than 512 pixels on the shortest side) and highly compressed images ($(\text{Height} \times \text{Width}) / \text{Filesize} < 0.15$).
(2) \textbf{Aesthetics filtering} discards images with ArtiMuse~\citep{cao2025artimusefinegrainedimageaesthetics} scores below 60.
(3) \textbf{Quality filtering} eliminates images with DeQA-Score~\citep{deqa_score} under 4.0.
After filtering, 140 million high-quality images are retained.

\paragraph{Image Captioning.} 
We generate captions using Qwen3-VL-235B-22B.
To retain real-world knowledge, the VLM evaluates the original web text: if informative, it incorporates this information into the caption (e.g., using ``Corgi Dog'' instead of a generic ``dog''), producing richer and more accurate descriptions.

\subsection{Image Editing}
\paragraph{Data Source.} 
Our image editing data combines open-source datasets and synthesized pairs.
We incorporate X2Edit~\citep{ma2026x2edit}, OmniEdit~\citep{wei2024omniedit}, Nano-consistent-150k~\citep{ye2025echo}, Pico-Banana~\citep{qian2025pico}, UniWorld~\citep{lin2025uniworld}, StructVisuals~\citep{zhuo2025factuality}, UnicEdit~\citep{ye2025unicedit}, and CrispEdit~\citep{chow2025editmgt}.
We also synthesize high-fidelity editing pairs by processing images from our generation dataset through an automated pipeline, further expanding data diversity while ensuring consistency between generation and editing tasks.

\paragraph{Instruction Refinement.}
We use Qwen3-VL-235B-22B for quality control.
First, we filter out ``failed'' samples where editing produces no observable change or introduces visual artifacts. 
Second, for high-quality transformations with inaccurate or vague instructions, the VLM rewrites instructions based on actual visual changes.
This ensures both visual integrity and precise instruction-image alignment.

\subsection{Interleaved Data}
\paragraph{Data Source \& Filtering.} 
We construct interleaved image-text data from the Koala36M~\citep{wang2025koala} video corpus through strict filtering:
(1) Duration filtering discards clips longer than 30 seconds or shorter than 10 seconds to minimize fragmentation errors.
(2) Quality filtering retains the top 50\% by aesthetic score ($>$ 4.0) and clarity ($>$ 0.7).
(3) Motion filtering requires a motion score greater than 4 to avoid degenerate solutions where the model generates static images.
This pipeline removes approximately 75\% of raw data, yielding 6M refined clips free from blur, static scenes, and low-aesthetic content.
We sample frames every 5 seconds, producing interleaved sequences of 2--6 frames.

\paragraph{Interleaved Captioning.} 
We use Qwen3-VL-235B-A22B to generate detailed descriptions of actions and scene changes from frame sequences.
Additionally, we generate user instructions tailored to these sequences, providing high-quality instruction-following data for SFT.

\subsection{Reasoning-Augmented Data}
To equip LLaDA2.0-Uni with reasoning capabilities, we incorporate a dedicated dataset comprising two components: reasoning-based image generation and interleaved reasoning.
We source this data from Flux-6M~\citep{fang2025flux}, Zebra-CoT~\citep{li2026zebracot}, and Weave~\citep{weave2024}, totaling approximately 8M samples for SFT.
This data enables chain-of-thought reasoning before image generation and multi-step reasoning across interleaved image-text sequences.

\section{Model Training}
\subsection{Training Recipe}
Our training pipeline consists of three stages that progressively enhance model capabilities: foundational cross-modal alignment, multi-task pre-training, and supervised fine-tuning.
Table~\ref{tab:foundational-stages} summarizes the data composition, token scale, and training configurations for each stage.

\begin{table}[htbp]
\centering
\small 
\caption{Overview of the Training Stages. This includes details on data composition, token scale, sequence length, and trainable components.}
\label{tab:foundational-stages}
\renewcommand{\arraystretch}{1.0}
\resizebox{\linewidth}{!}{
\begin{tabular}{cccc}
\toprule
\textbf{Stages\&Objective} & \textbf{S0: Vision-Language Alignment} & \textbf{S1: Multi-task Pre-training} & \textbf{S2: Supervised Fine-Tuning}  \\
\midrule
\textbf{\makecell[c]{Understanding\\Data}} & \makecell[c]{Image Caption, Text} & \makecell[c]{Image Caption, Text, \\ OCR, Grounding, \\ Counting, Video Data,\\ Multimodal VQA} & \makecell[c]{High-quality Multimodal VQA\\ High-quality Text QA \\ Interleaved Reasoning} \\
\midrule
\textbf{\makecell[c]{Generation\\Data}} & Text-to-image & \makecell[c]{Text-to-image \\ Image Editing \\ Interleaved Generation}& \makecell[c]{High-quality Image Generation \\ Image Generation with CoT\\ High-quality Image Editing\\ High-quality Interleaved Generation\\ Interleaved Reasoning}  \\
\midrule
\textbf{Gen. Resolution}& $256\rightarrow512$ & 512& 512~~(diffusion decoder $\rightarrow$ 1024)  \\
\midrule
\textbf{Under. Max Edge}& 800 & 800& 800  \\
\midrule
\textbf{Training Tokens} & 100B & 210B & 80B \\
\midrule
\textbf{Sequence length} & 8192 & 8192& $8192\rightarrow16384$\\
\bottomrule
\end{tabular}
}
\vspace{-0.15in}
\end{table}

\noindent
\textbf{Stage 0: Vision–Language Alignment.} 
The primary objective of stage 0 (S0) is to align visual and linguistic representations within the dLLM backbone.
We use high-quality image–caption pairs and visual knowledge datasets, supplemented with pure text data to preserve language capabilities.
During training, a random masking strategy is applied to a subset of text and image tokens: for generation tasks, only image tokens are masked; for understanding tasks, only text tokens are masked.
To handle long visual token sequences, we adopt a progressive arbitrary resolution scheme: generation starts at $256 \times 256$ ($\sim$256 tokens) and transitions to $512 \times 512$ ($\sim$1024 tokens), while understanding consistently uses $800 \times 800$ with arbitrary resolution ($\sim$2048 tokens).

\noindent
\textbf{Stage 1: Multi-task Pre-training.} 
The model is trained on diverse multimodal data to develop comprehensive understanding and generation capabilities.
Visual understanding data includes image–text interleaved data, OCR, and visual counting/grounding tasks. 
Generation data includes image editing, subject-driven generation, controllable generation, style transfer using reference images, and multi-view generation tasks. 
This stage strengthens cross-modal connections and enables the model to handle increasingly complex tasks.

\noindent
\textbf{Stage 2: Supervised Fine-tuning.}
The Supervised Fine-Tuning (SFT) process is conducted in two stages: an initial phase at 8k context length for fundamental instruction-following capabilities, followed by expansion to 16k context for complex visual reasoning and generation.

\subsection{Pre-Training Optimization}
We adopt the Block Diffusion Language Model (BDLM)~\citep{arriola2025block} training objective, which extends standard discrete diffusion by operating on block-level masked regions rather than individual tokens.
This design enables parallel decoding while maintaining coherent context within each block, making it well-suited for the variable-length sequences common in multimodal tasks.

\noindent
\textbf{BDLM Loss.}
The training loss under the BDLM paradigm is defined as:
\begin{equation}\label{eq:bdlm}
    \mathcal{L}_{\text{BDLM}}(\theta) = - \mathbb{E}_{t, \bm{x}_0, \bm{x_t}} \left[ \frac{\alpha_{t}'}{1-\alpha_{t}}\sum_{k=1}^{K} \sum_{i=1}^{L_B} \mathbb{1}[x_{t,k}^i=\text{[MASK]}] \log p_{\theta}(\bm{x}_{0,k}^i | \bm{x}_{0,<k}, \bm{x}_{t,k}) \right],
\end{equation}
where the expectation is over timestep $t$, the clean sequence $\bm{x}_0$, and its corrupted version $\bm{x}_t$ (tokens masked with probability $1 - \alpha_t$). 
The indicator $\mathbb{1}[\cdot]$ ensures predictions are made only for masked tokens, and $-\alpha'_t / (1 - \alpha_t)$ is the diffusion-derived time weight. 
We define: $K = L_{\text{total}} / L_B$ as the number of blocks, $L_B$ as the block size, $x^i_{t,k}$ as the $i$-th token in block $k$, $\bm{x}_{0,<k}$ as the preceding clean blocks, and $\bm{x}_{t,k}$ as the noisy version of the current block.
\looseness=-1

\noindent
\textbf{Load Balancing Strategy.}
In MoE models, imbalanced expert utilization can lead to routing collapse.
We adopt an auxiliary-loss-free load balancing mechanism~\citep{deepseekv3} that promotes differentiated expert specialization while encouraging uniform workload distribution.
To improve numerical stability, we scale routing gate outputs by a factor of 2.5, stabilizing their root-mean-square (RMS) magnitude.
The auxiliary-loss-free bias is updated according to~\citep{sjl_moe}:
\begin{equation}
b_{i}=b_{i}+u\times \frac{(F_{i}-Q_{i})}{\sqrt{\frac{1}{n}\sum_{j=1}^{n}{(F_{j}-Q_{j})^2}}} \,,
\end{equation}
where $F=\mathbb{E}(f)$ denotes the current expert load distribution induced by the bias $b$, and $Q=[\frac{1}{n},\frac{1}{n},\dots,\frac{1}{n}]$ represents the ideal uniform distribution over $n$ experts. 
This RMSNorm-style normalization smooths bias updates, leading to stable load balancing throughout training.

\subsection{Supervised Fine-Tuning Optimization}
During SFT, we adopt the same load balancing strategy while introducing complementary masking and a mask token reweighting loss to handle variable-length sequences.

\noindent
\textbf{Mask Token Reweighting Loss.}
We adapt the BDLM objective to be conditional on an input prompt $c$:
\begin{equation}\label{eq:bdlm}
    \mathcal{L}_{\text{SFT}}(\theta) = - \mathbb{E}_{t, (c,\bm{x}_0), \bm{x_t}} \left[ \frac{\alpha_{t}'}{1-\alpha_{t}}\sum_{k=1}^{K} \sum_{i=1}^{L_B} \mathbb{1}[x_{t,k}^i=\text{[MASK]}] \log p_{\theta}(\bm{x}_{0,k}^i | c,\bm{x}_{0,<k}, \bm{x}_{t,k}) \right] \,.
\end{equation}

A key challenge in SFT is that sample lengths vary significantly---by up to two orders of magnitude.
Naive token-averaged loss causes gradients to be dominated by long sequences, while sample-level averaging encourages brevity.
We therefore propose a re-weighting mechanism to balance these extremes:
\begin{equation}
    \mathcal{L}_{\text{MTRS}} = \frac{\sum_{j} \beta_j \mathcal{L}_{\text{SFT}}^{(j)}}{\sum_{j} \beta_j} \,, \quad \text{where} \quad \beta_j = \frac{1}{\sqrt{\sum_{k=1}^{K} \sum_{i=1}^{L_B} \mathbb{1}[x_{t,k}^{i,(j)}=\text{[MASK]}]}} \,.
\end{equation}
The scaling factor $\beta_j$ is the inverse square root of the number of masked tokens in sample $j$, equilibrating gradient contributions across diverse response lengths.

\noindent
\textbf{Complementary Masking.}
Complementary masking~\citep{li2025lavida} is a strategy to enhance data efficiency for dLLMs by constructing two antithetical training instances from a single sequence $x_0$: a primary noised sequence $x_t$ and a complementary sequence $x'_t$ using the inverse mask.
This design ensures that every token position appears uncorrupted exactly once per pair, thereby doubling effective information utilization and eliminating token-level sampling bias.
We adapt this strategy in our framework.

\subsection{Diffusion Decoder Training}
\label{sec:decoder_training}
\paragraph{Training Paradigm.} 
We optimize the diffusion decoder via the standard flow matching objective~\citep{lipman2022flow}. The overall training trajectory is decoupled into a preliminary warm-up followed by a progressive two-stage fine-tuning scheme. The flow matching loss is formulated as:
\begin{equation}
    \mathcal{L}_{\text{FM}}(\theta) = \mathbb{E}_{\bm{x}_0, \bm{x}_1, \bm{z}, t} \left[ \| \bm{v}_{\theta,t}(\bm{x}_t, \bm{z}) - \bm{v}_{t} \|^2_2 \right],
\end{equation}
where $\bm{z}$ represents the conditioned semantic visual tokens, $\bm{v}_{\theta,t}$ denotes the velocity field predicted by the network at timestep $t$, and $\bm{v}_{t}$ is the target velocity. 
The overall training process is divided into three stages:
\begin{itemize}[leftmargin=16pt]
    \item \textbf{Stage 1: Warm-up.} We freeze the semantic processor and update only the remaining modules to establish cross-modal alignment while preserving pre-trained priors.

    \item \textbf{Stage 2: Multi-domain Generalization.} Following the warm-up, we unfreeze all parameters and
    fine-tune on diverse domains for robust generalization.

    \item \textbf{Stage 3: High-fidelity Refinement.} 
    In the final stage, we refine on high-quality data to elevate aesthetic fidelity and fine-grained visual details.
\end{itemize}

\paragraph{Few-step Generation.}
To accelerate visual generation, we adopt a lightweight consistency-based distillation framework~\citep{sun2026duality} for the diffusion decoder.
This method requires only an auxiliary projection layer---a final additional layer added to the decoder backbone---which will be discarded at inference time.
The distillation objective combines flow matching with a consistency term:
\looseness=-1
\begin{equation}
    \mathcal{L}_{\text{Distill}}(\theta) = 
    \mathbb{E}_{\bm{x}_0, \bm{z}, t} \left[ \| \bm{v}_{\theta,t} - \bm{v}_{t} \|^2_2 + \| \bm{u}_{\theta,t} - \bm{v}_{t} + t \cdot \frac{\text{d}\bm{u}_{\theta^-,t}}{\text{d}t} \|^2_2 \right], \text{where} \,\, \bm{u}_{\theta^-,t}=\text{stop\_grad(}\bm{u}_{\theta,t}\text{)} \,,
    \label{distill_loss}
\end{equation}
where $\bm{v}_{t}$ is the target velocity, $\bm{v}_{\theta,t}, \bm{u}_{\theta,t}$ are the dual outputs of the diffusion decoder.
The time derivative $\nicefrac{\text{d}\bm{u}_{\theta^-,t}}{\text{d}t}$ is a Jacobian-vector product (JVP)~\citep{lu2024simplifying} output of the diffusion decoder, where the JVP calculation is approximated using the second-order difference technique proposed in UCGM~\citep{sun2025unified}.
This strategy enables 8-step CFG-free inference while maintaining high image quality.

\begin{figure}[htb]
    \centering
    \vspace{-0.15in}\includegraphics[width=0.85\linewidth]{figs/packing.pdf}
    \vspace{-0.05in}
    \caption{
    Data Packing Strategy for Efficient Training. Multiple shorter samples are concatenated into fixed-length sequences, minimizing padding tokens and improving GPU utilization.
    }
    \label{fig:pack}
    \vspace{-0.1in}
\end{figure}
\subsection{Infrastructure for Training Efficiency}
\noindent
\textbf{Image Tokens Pre-extraction.} 
LLaDA2.0-Uni employs a Vector Quantized (VQ) tokenizer to transform images into discrete visual tokens, incurring substantial computational cost during training.
We adopt an offline pre-extraction strategy: prior to training, the entire dataset is processed through the frozen tokenizer, with token indices stored on disk.
During training, the data loader retrieves pre-extracted tokens directly, eliminating repeated encoder passes and significantly accelerating the pipeline.

\noindent
\textbf{Load Balancing via Data Packing.}
In LLaDA2.0-Uni, sequence lengths vary significantly across multimodal tasks (e.g., short text-only tasks versus long image-generation tasks).
Traditional batching strategy requires extensive padding to match the longest sequence, wasting computation on padding tokens.
We address this with an offline data packing strategy that consolidates multiple shorter samples into fixed-length sequences (Figure~\ref{fig:pack}), significantly increasing effective token throughput for both pre-training and post-training.

\noindent
\textbf{Distributed Framework.}
We employ dFactory~\citep{dFactory} as the primary training engine for both pre-training and post-training phases---a high-efficiency framework specifically optimized for Diffusion Large Language Models. 
Built upon the VeOmni~\citep{ma2025veomni} distributed training ecosystem, dFactory enables the flexible deployment of sophisticated parallelization strategies. 
\section{Experiments}
\subsection{Multimodal Understanding}
\subsubsection{Evaluation Settings}
\paragraph{Benchmarks.} We evaluate LLaDA2.0-Uni across 21 multimodal understanding benchmarks, focusing on three core capabilities: general VQA, reasoning, and OCR/document understanding:
\begin{itemize}[leftmargin=16pt]
    \item \textbf{General Tasks:}
    MMStar~\citep{mmstar}, MMBench~\citep{Mmbench}, MME~\citep{fu2023mme}, HallusionBench~\citep{Hallusionbench},  RealWorldQA~\citep{RealworldQA} and SimpleVQA~\citep{Simplevqa}.
    \item \textbf{Reasoning Tasks:} 
    MMMU~\citep{yue2024mmmu}, and MMMU-Pro~\citep{yue2025mmmupro}, MathVista~\citep{mathvista}, We-Math~\citep{We-math}, MathVision~\citep{wang2024measuring}, and MathVerse~\citep{zhang2024mathverse}.
    \item \textbf{OCR\&Chart Tasks:} 
    ChartQA~\citep{Chartqa}, DocVQA~\citep{Docvqa}, InfoVQA~\citep{InfoVQA}, CharXiv~\citep{Charxiv}, OCRBench~\citep{Ocrbench}, and AI2D~\citep{AI2D}.
    \item \textbf{Other Multimodal Tasks:}
    CountBench~\citep{countbench}, VLRewardBench~\citep{VLRewardBench}, and V$^*$~\citep{vstar}.
\end{itemize}

\paragraph{Baselines.} 
To evaluate the multimodal performance of LLaDA2.0-Uni, we compare it against an extensive set of baselines.
We first compare LLaDA2.0-Uni with leading specialized VLMs, including Qwen2.5-VL-7B~\citep{bai2025qwen2} (AR-based) and LLaDA-V~\citep{you2025llada} (diffusion-based).
Furthermore, we evaluate LLaDA2.0-Uni against state-of-the-art unified models categorized into: 1) AR-based models, such as BAGEL~\citep{bagel} and InternVL-U~\citep{tian2026internvl}; and 2) diffusion-based models, such as Lumina-DiMOO~\citep{dimoo} and LLaDA-o~\citep{you2026lladao}.

\subsubsection{Multimodal Understanding Performance}
As shown in Table~\ref{tab:understanding}, LLaDA2.0-Uni demonstrates strong and comprehensive multimodal understanding capabilities.
Compared to existing diffusion-based unified models like Lumina-DiMOO and LLaDA-o, LLaDA2.0-Uni achieves significant improvements across all major categories, particularly in general VQA tasks (e.g., MMStar: 64.1 vs. 58.0) and complex reasoning tasks (e.g., MMMU: 50.1 vs. 44.9).
Furthermore, LLaDA2.0-Uni also delivers consistently high performance in challenging OCR and document understanding scenarios, where baselines like Lumina-DiMOO struggle significantly.
Most impressively, LLaDA2.0-Uni performs on par with state-of-the-art specialized VLMs such as Qwen2.5-VL-7B, even slightly outperforming it on specific metrics like MMStar (64.1 vs. 63.9) and CountBench (86.0 vs. 84.9).
Overall, these results confirm that LLaDA2.0-Uni closes the gap between unified diffusion architectures and top-tier specialized VLMs.

\begin{table}[t]
\centering
\renewcommand{\arraystretch}{1.2}
\setlength\tabcolsep{2pt}
\small
\caption{The Overall Comparison of LLaDA2.0-Uni and Existing Specialist VLMs and Unified Models.}
\label{tab:exp-overall}

\definecolor{myblue}{rgb}{0.9, 0.94, 1.0}

\begin{tabularx}{\textwidth}{l *{6}{>{\centering\arraybackslash}X} >{\columncolor{myblue}\centering\arraybackslash}X}
\toprule
& \multicolumn{2}{c}{\textbf{Specialist VLMs}} & \multicolumn{5}{c}{\textbf{Unified Models}} \\
\cmidrule(lr){2-3} \cmidrule(lr){4-8}
& \textbf{Qwen2.5-VL-7B} & \multirow{2}{*}{\textbf{LLaDA-V}} & \multirow{2}{*}{\textbf{BAGEL}} & \multirow{2}{*}{\textbf{InternVL-U}} & \textbf{Lumina-DiMOO} & \multirow{2}{*}{\textbf{LLaDA-o}} & \textbf{LLaDA2.0-Uni} \\

\midrule
\multicolumn{8}{l}{\textbf{General Tasks}} \\
\midrule
MMStar & 63.9 & 60.1 & 67.0 & 54.7 & 61.0 & 58.0 & 64.1 \\
MMBench$_{\text{EN}}$& 83.5 & 82.9 & 85.0 & 75.3 & 84.5 & 71.1 & 81.5 \\
MMBench$_{\text{CN}}$ & 83.4 & 70.1 & 82.4 & 73.6 & 71.8 & 69.9 & 81.2 \\
MME-C & 62.4 &49.1 & 66.7 & 27.9 & 35.2 & 52.7 & 58.7 \\
HallusionBench & 51.9 &39.2 & 52.5 & 44.8 & 32.9 & 47.4 & 50.2 \\
RealWorldQA & 68.5 & 63.2 & 73.9 & 56.4 & 52.4 & 60.8 & 66.7 \\
SimpleVQA & 47.9 & 26.0 & 41.9 & 20.7 & 12.1 & 29.2 & 44.0 \\

\midrule
\multicolumn{8}{l}{\textbf{Reasoning Tasks}} \\
\midrule
MMMU$_{\text{val}}$ & 51.3 & 48.6 & 55.3 & 54.7 & 58.6 & 44.9 & 50.1 \\
MMMUPro$_{\text{standard}}$ & 38.3 & 35.2 & 37.1 & 20.8 & 20.6 & 28.3 & 34.0 \\
MathVista$_{\text{mini}}$ & 68.2 & 59.7 & 73.1 & 55.8 & 10.3 & 66.1 & 68.1 \\
MathVision$_{\text{mini}}$ & 22.4 & 21.2 & 24.1 & 22.1 & 13.1 & 15.7 & 26.7 \\
WeMath & 33.3 & 24.6 & 45.8 & 18.3 & 6.2 & 29.3 & 29.3 \\

\midrule
\multicolumn{8}{l}{\textbf{OCR \& Chart Tasks}} \\
\midrule
CharXiv(DQ) & 73.9 & 47.0 & 70.6 & 53.3 & 27.8 & 69.8 & 68.4 \\
ChartQA & 84.1 & 78.3 & 74.3 & 76.6 & 8.3 & 87.9 & 80.1 \\
OCRBench & 84.2 & 63.2 & 73.3 & 83.9 & 7.6 & 74.6 & 75.7 \\
DocVQA & 94.9 & 83.9 & 94.3 & 85.4 & 7.2 & 91.5 & 89.5 \\
$\mathrm{AI2D}_{\text{w mask}}$ & 82.6 & 77.8 & 88.9 & 76.3 & 43.2 & 79.3 & 82.0 \\
InfoVQA & 80.3 & 66.3 & 60.7 & 68.3 & 6.2 & 54.7 & 70.1\\

\midrule
\multicolumn{8}{l}{\textbf{Other Tasks}} \\
\midrule
CountBench & 84.9 & 75.1 & 93.2 & 62.2 & 48.4 & 91.7 & 86.0 \\
VL-RewardBench & 45.2 & 46.0 & 28.9 & 46.4 & 51.7 & 42.4 & 47.8 \\
V$^{*}$ & 80.1 & 41.8 & 67.5 & 51.3 & 52.7 & 57.6 & 61.8 \\

\bottomrule
\end{tabularx}
\label{tab:understanding}
\vspace{-0.15in}
\end{table}

\subsection{Text-to-Image Generation}
\subsubsection{Evaluation Settings}
\paragraph{Benchmarks.} 
We conduct a comprehensive evaluation using a suite of established public benchmarks, including GenEval~\citep{ghosh2023geneval}, DPG-Bench~\citep{hu2024ella}, One-IG Bench~\cite{chang2025oneig}, and UniGenBench~\cite{UniGenBench++} for general generative capabilities, as well as CVTG-2K~\citep{du2025textcrafter} for text rendering proficiency. 
To further evaluate reasoning-informed image generation, we also test LLaDA2.0-Uni on the WISE-Bench~\citep{niu2025wise}. 

\paragraph{Baselines.} To comprehensively assess the text-to-image generation capabilities of LLaDA2.0-Uni, we benchmark our model against a diverse spectrum of strong baselines.
We first compare LLaDA2.0-Uni with leading specialized generation models (Gen. Only), including diffusion models like FLUX.1 [Dev]~\citep{flux}, Lumina-Image 2.0~\citep{qin2025lumina}, Seedream 3.0~\citep{gao2025seedream}, Qwen-Image~\citep{wu2025qwen}, LongCat-Image~\citep{team2025longcat}, and Z-Image~\citep{cai2025z}, as well as AR-based models like Emu3~\citep{wang2024emu3} and Lumina-mGPT 2.0~\citep{xin2025lumina}.
Furthermore, we evaluate LLaDA2.0-Uni against state-of-the-art unified models categorized into: 
1) AR-based and hybrid (AR + Diff.) models, including Janus-Pro~\citep{Janus}, BAGEL~\citep{bagel}, and OmniGen2~\citep{wu2025omnigen2}, Hunyuan Image 3.0~\citep{cao2025hunyuanimage}, NextFlow~\citep{zhang2026nextflow}, and InternVL-U~\citep{tian2026internvl}; 
and 2) discrete diffusion (D-Diff.) and hybrid diffusion (D-Diff. + Diff.) models, including MMaDA~\citep{Mmada}, Lumina-DiMOO~\citep{dimoo} and LLaDA-o~\citep{you2026lladao}.

\begin{table}[!tb]
\centering
\renewcommand{\arraystretch}{1.2}
\setlength\tabcolsep{1.5pt} 
\small
\newcolumntype{C}{>{\centering\arraybackslash}X}

\caption{Comparison of Text-to-Image Generation Ability on GenEval Benchmark.}
\begin{tabularx}{\linewidth}{ccc| *{6}{C}|c} 
\toprule
\textbf{Type} & \textbf{Model} &\textbf{Arch.} &\textbf{\makecell{Single\\Object}} & \textbf{\makecell{Two\\Object}} & \textbf{Counting} & \textbf{Colors} & \textbf{Position} & \textbf{\makecell{Attribute\\Binding}}& \textbf{~Overall$\uparrow$} \\
\midrule
\multirow{7}{*}{\rotatebox[origin=c]{90}{Gen. Only}} 
& FLUX.1 [Dev] & Diff. & 0.98& 0.81 & 0.74& 0.79& 0.22& 0.45 &0.66 \\
& Emu3-Gen & AR & 0.98 & 0.71 & 0.34 & 0.81 & 0.17 & 0.21 & 0.54 \\
& Lumina-mGPT 2.0 &AR &0.99 &0.87 &0.44 &0.85 &0.44 &0.54 &0.69\\
& Seedream 3.0 &Diff. &0.99 &0.96 &0.91 &0.93 &0.47 &0.80 &0.84 \\
&Qwen-Image &Diff. &0.99 &0.92 &0.89 &0.88 &0.76 &0.77 &0.87 \\
&LongCat-Image &Diff. &0.99 &0.98 &0.86 &0.86 &0.75 &0.73 &0.87\\
&Z-Image-Turbo &Diff. &1.00 &0.95 &0.77 &0.89 &0.65 &0.68 &0.82 \\

\midrule
\multirow{10}{*}{\rotatebox[origin=c]{90}{Unified}} 
& Janus-Pro & AR & 0.99 & 0.89 & 0.59 & 0.90 & 0.79 & 0.66 & 0.80 \\
& BAGEL & AR + Diff. & 0.99 & 0.94 & 0.81 & 0.88 & 0.64 & 0.63 & 0.82 \\
& OmniGen2 & AR + Diff. &1.00 &0.95 &0.64 &0.88 &0.55 &0.76 &0.80 \\
& HunyuanImage-3.0 & AR + Diff. &1.00 &0.92 &0.48 &0.82 &0.42 &0.63 &0.72\\
& NextFlow & AR + Diff. &0.98 &0.92 &0.73 &0.90 &0.77 &0.69 &0.83 \\
& InternVL-U & AR + Diff. & 0.99 & 0.94 & 0.74 & 0.91 & 0.77 & 0.74 & 0.85 \\
& MMaDA & D-Diff. & 0.99 & 0.76 & 0.61 & 0.84 & 0.20 & 0.37 & 0.63 \\
& Lumina-DiMOO & D-Diff. &1.00 &0.94 &0.85 &0.89 &0.85 &0.76 &0.88 \\
& LLaDA-o & D-Diff. + Diff. &0.99 &0.98 &0.73 &0.96 &0.69 &0.83 &0.86 \\
\rowcolor{myblue}
& LLaDA2.0-Uni & D-Diff. + Diff. &1.00 & 0.98 & 0.73  & 0.92 & 0.90 & 0.84 & 0.89 \\
\bottomrule
\end{tabularx}
\label{table:geneval}
\vspace{-0.06in}
\end{table}

\begin{table}[!tb]
\centering
\renewcommand{\arraystretch}{1.2}
\setlength\tabcolsep{1pt} 
\small
\newcolumntype{C}{>{\centering\arraybackslash}X}

\caption{Comparison of Text-to-Image Generation Ability on DPG Benchmark.}
\begin{tabularx}{\linewidth}{ccc| *{5}{C}|c} 
\toprule
\textbf{Type} & \textbf{Model} &\textbf{Arch.} & \textbf{Global} & \textbf{Entity} & \textbf{ Attribute} & \textbf{Relation} & \textbf{Other} & \textbf{~Overall$\uparrow$} \\
\midrule
\multirow{7}{*}{\rotatebox[origin=c]{90}{Gen. Only}} 
& FLUX.1 [Dev] & Diff. & 74.35 & 90.00 & 88.96 & 90.87 & 88.33 & 83.84 \\
& Emu3-Gen & AR & 85.21 & 86.68 & 86.84 & 90.22 & 83.15 & 80.60 \\
& Lumina-Image 2.0 & Diff. & 86.63 & 91.97 & 90.20 & 94.85 & 84.80 & 87.20 \\
& Seedream 3.0 & Diff. &94.31 &92.65 &91.36 &92.78 &88.24 &88.27 \\
& Qwen-Image & Diff. &91.32 &91.56 &92.02 &94.31 &92.73 &88.32 \\
&LongCat-Image &Diff. &89.10 &92.54 &92.00 &93.28 &87.50 &86.80\\
& Z-Image-Turbo & Diff. &91.29 &89.59 &90.14 &92.16 &88.68 &84.86 \\

\midrule
\multirow{10}{*}{\rotatebox[origin=c]{90}{Unified}} 
& Janus-Pro & AR & 86.90 & 88.90 & 89.40 & 89.32 & 89.48 & 84.19 \\
& BAGEL & AR + Diff. & 88.94 & 90.37 & 91.29 & 90.82 & 88.67 & 85.07 \\
& OmniGen2 & AR + Diff. & 88.81 & 88.83 & 90.18 & 89.37 & 90.27 & 83.57 \\
& HunyuanImage-3.0 & AR + Diff. &92.12 &92.53 &89.13 &92.13 &91.92 &86.10\\
& NextFlow & AR + Diff. & 92.40 &90.05 &90.51 &92.72 &91.14 & 86.00 \\
& InternVL-U & AR + Diff. & 90.39 & 90.78 & 90.68 & 90.29 & 88.77 &85.18 \\
& MMaDA & D-Diff. & 77.81 & 78.48 & 81.74 & 84.79 & 63.20 & 69.97 \\
& Lumina-DiMOO & D-Diff. & 81.46 & 92.08 & 88.98 & 94.31 & 82.00 & 86.04 \\
& LLaDA-o &D-Diff. + Diff. & 92.91 & 93.30 & 90.40 & 91.75 &92.79 & 87.04 \\
\rowcolor{myblue}
& LLaDA2.0-Uni &D-Diff. + Diff. & 91.14 & 93.55 & 91.98 & 92.17 & 93.18 & 87.76 \\
\bottomrule
\end{tabularx}
\label{table:dpg_benchmark}
\vspace{-0.15in}
\end{table}

\begin{table}[!tb]
\centering
\renewcommand{\arraystretch}{1.2}
\setlength\tabcolsep{3pt}
\small
\newcolumntype{C}{>{\centering\arraybackslash}X}

\caption{Comparison of Text-to-Image Generation Ability on OneIG-EN Benchmark.}
\begin{tabularx}{\linewidth}{ccc| *{5}{C}|c} 
\toprule
\textbf{Type} & \textbf{Model} & \textbf{Arch.} & \textbf{Alignment} &\textbf{Text} &\textbf{Reasoning} &\textbf{Style} &\textbf{Diversity} &\textbf{~Overall$\uparrow$} \\
\midrule
\multirow{5}{*}{\rotatebox[origin=c]{90}{Gen. Only}} 
& FLUX.1 [Dev] & Diff. & 0.786 & 0.523 & 0.253 & 0.368 & 0.238 & 0.434 \\
& Lumina-Image 2.0 & Diff. & 0.819 &0.106 &0.270 & 0.354 &0.216 & 0.353 \\
& Seedream 3.0 & Diff. &0.818 &0.865 &0.275 &0.413 &0.277 &0.530 \\
& Qwen-Image & Diff. &0.882 &0.891 &0.306 &0.418 &0.197 &0.539 \\
& Z-Image-Turbo & Diff. &0.840 &0.994 &0.298 &0.368 &0.139 &0.528 \\

\midrule
\multirow{6}{*}{\rotatebox[origin=c]{90}{Unified}} 
& Janus-Pro & AR & 0.553 & 0.001 & 0.139 & 0.276 & 0.365 & 0.267 \\
& BAGEL & AR + Diff. & 0.769 & 0.244 & 0.173 & 0.367 & 0.251 & 0.361 \\
& OmniGen2 & AR + Diff. & 0.804 &0.680 &0.271 &0.377 &0.242 &0.475 \\
& InternVL-U & AR + Diff. &0.820 &0.740 &0.270 &0.400 &0.250 &0.500\\
& Lumina-DiMOO & Diff. &0.820& 0.550& 0.280& 0.400& 0.230 & 0.460  \\
\rowcolor{myblue}
&LLaDA2.0-Uni &D-Diff. + Diff. &0.882 & 0.661 &0.323 &0.400 &0.259 &0.505 \\
\bottomrule
\end{tabularx}
\label{table:oneig_en}
\vspace{-0.06in}
\end{table}

\begin{table}[!tb]
\centering
\renewcommand{\arraystretch}{1.2}
\setlength\tabcolsep{1pt}
\small
\newcolumntype{C}{>{\centering\arraybackslash}X}

\caption{Comparison of Text-to-Image Generation Ability on UniGenBench.}
\begin{tabularx}{\linewidth}{ccc|*{10}{C}|c} 
\toprule
\textbf{Type} & \textbf{Model} & \textbf{Arch.} & \textbf{Style} & \textbf{World} & \textbf{Attr.} & \textbf{Action} & \textbf{Relat.} & \textbf{Logic} & \textbf{Gram.} & \textbf{Comp.} & \textbf{Layout} & \textbf{Text} & \textbf{~Overall$\uparrow$} \\
\midrule
\multirow{5}{*}{\rotatebox[origin=c]{90}{Gen. Only}} 
&FLUX.1 [dev] & Diff. & 83.90 & 88.92 & 67.84 & 62.17 & 67.26 & 30.91 & 60.96 & 47.04 & 71.83 & 32.18 &61.30   \\

&Emu3-Gen & AR   & 86.80 & 77.06 & 51.39 & 40.11 & 49.75 & 19.32 & 52.94 & 36.86 & 44.78 &  1.15 &46.02 \\

&Seedream 3.0 &Diff. &98.19 &94.90 &84.62 &83.14 &80.18 &51.83 &60.30 &72.32 &88.74 &69.86 &78.41\\

&Qwen-Image & Diff. &94.70 &94.15 &87.93 &82.60 &80.08 &51.59 &60.96 &72.94 &86.57 &72.13 &78.36\\

&Z-Image & Diff. &96.80 &94.46 &82.48 &78.90 &80.20 &49.08 &68.98 &76.80 &84.89 &68.39 &78.10\\

\midrule
\multirow{6}{*}{\rotatebox[origin=c]{90}{Unified}} 
&Janus-Pro & AR & 90.80 & 86.71 & 67.74 & 64.26 & 68.40 & 37.05 & 64.44 & 62.11 & 72.01 &  2.59 &61.61   \\

&BAGEL &AR + Diff. & 90.20 & 85.60 & 67.74 & 61.98 & 70.69 & 30.23 & 66.44 & 58.12 & 76.49 &  7.76 &  61.53  \\

&OmniGen2 & AR + Diff. &91.90 &86.39 &72.12 &62.83 &68.27 &32.50 &59.89 &56.31 &71.64 &29.02 &63.09\\

&MMaDA & D-Diff. &82.40 &56.65 &48.39 &37.83 &50.25 &17.95 &55.75 &32.35 &30.22 &1.15 &41.35\\

&Lumina-DiMOO & D-Diff. &89.70 &90.03 &81.62 &71.12 &78.43 &45.45 &70.45 &73.32 &82.84 &25.57 &71.12\\

\rowcolor{myblue}
&LLaDA2.0-Uni & D-Diff. + Diff. &95.30  &93.67 &91.77 &85.65 &86.42 &63.99 &72.19 &85.82 &90.30 &31.23 &79.63\\
\bottomrule
\end{tabularx}
\label{table:unigen}
\vspace{-0.08in}
\end{table}

\begin{table}[!tb]
\centering
\renewcommand{\arraystretch}{1.2}
\setlength\tabcolsep{2pt}
\small
\newcolumntype{C}{>{\centering\arraybackslash}X}

\caption{Comparison of Text Rendering Ability on CVTG-2K Benchmark.}
\begin{tabularx}{\linewidth}{ccc| *{5}{C}| CC} 
\toprule
\multirow{2}{*}{\textbf{Type}} & \multirow{2}{*}{\textbf{Model}} &\multirow{2}{*}{\textbf{Arch.}} & \multicolumn{5}{c|}{\textbf{Word Accuracy~$\uparrow$}} & \multirow{2}{*}{\textbf{NED~$\uparrow$}} & \multirow{2}{*}{\textbf{CLIPScore~$\uparrow$}} \\
\cmidrule(lr){4-8}
& & & 2 regions & 3 regions & 4 regions & 5 regions & average & & \\
\midrule
\multirow{5}{*}{\rotatebox[origin=c]{90}{Gen. Only}} 
& FLUX.1 [Dev] & Diff. & 0.608 & 0.553 & 0.466 & 0.431 & 0.496 & 0.687 & 0.740 \\
& Seedream 3.0 & Diff. &0.628 &0.596 &0.604 &0.561 & 0.592 &0.853 &0.782\\
& Qwen-Image & Diff. & 0.837 &0.836 &0.831 &0.816 &0.829 &0.912 &0.802 \\
& LongCat-Image & Diff. &0.912 &0.873 &0.855 &0.831 &0.865 &0.936 &0.785 \\
& Z-Image-Turbo & Diff. &0.887 &0.866 &0.862 &0.834 &0.858 &0.928 &0.804\\
\midrule
\multirow{3}{*}{\rotatebox[origin=c]{90}{Unified}} 
& BAGEL & AR + Diff. &0.498 &0.391 &0.332 &0.291 &0.356 &0.657 &0.779 \\
& InternVL-U & AR + Diff. &0.729 &0.660 &0.618 &0.549 &0.623 &0.804 &0.816\\
& Lumina-DiMOO & D-Diff. & 0.723 &0.646 &0.571 &0.505 &0.590  &0.805 &0.831 \\
\rowcolor{myblue} 
& LLaDA2.0-Uni & D-Diff. + Diff. & 0.788  & 0.776 & 0.763 & 0.746  &0.765  & 0.911  & 0.818\\
\bottomrule
\end{tabularx}
\label{table:cvtg2k_rendering}
\vspace{-0.15in}
\end{table}

\begin{table}[!tb]
\centering
\renewcommand{\arraystretch}{1.2}
\setlength\tabcolsep{1pt}
\small
\newcolumntype{C}{>{\centering\arraybackslash}X}
\vspace{-0.1in}
\caption{Comparison of Reasoning-Informed Image Generation Ability on WISE Benchmark.}
\begin{tabularx}{\linewidth}{ccc| *{6}{C} |c} 
\toprule
\textbf{Type} & \textbf{Model} & \textbf{Arch.} & \textbf{Cultural} & \textbf{Time} & \textbf{Space} & \textbf{Biology} & \textbf{Physics} & \textbf{Chem.} & \textbf{~Overall$\uparrow$} \\
\midrule
\multirow{5}{*}{\rotatebox[origin=c]{90}{Gen. Only}} 
& SD3-Medium & Diff. &0.43 &0.50 &0.52 &0.41 &0.53 &0.33 &0.45 \\
& FLUX.1 [Dev] & Diff. & 0.48 & 0.58 & 0.62 & 0.42 & 0.51 & 0.35 & 0.50 \\
& Emu3-Gen & AR &0.34 &0.45 &0.48 &0.41 &0.45 &0.27 &0.39 \\
& Qwen-Image & Diff. &0.62 &0.63 &0.77 &0.57 &0.75 &0.40 &0.62 \\
& LongCat-Image &Diff. &0.66 &0.61 &0.72 &0.66 &0.72 &0.49 &0.65 \\

\midrule
\multirow{7}{*}{\rotatebox[origin=c]{90}{Unified}} 
& Janus-Pro & AR & 0.30 & 0.37 & 0.49 & 0.36 & 0.42 & 0.26 & 0.35 \\
& BAGEL & AR + Diff. & 0.44 & 0.55 & 0.68 & 0.44 & 0.60 & 0.39 & 0.52 \\
& NextFlow & AR + Diff. &0.62 &0.60 &0.70 &0.54 &0.58 &0.38 &0.59 \\
&InternVL-U & AR + Diff. &0.37 &0.51 &0.68 &0.39 &0.62 &0.39 &0.46 \\
& Lumina-DiMOO & D-Diff. & 0.35 & 0.43 &0.59 &0.31 &0.49 &0.34 &0.40 \\
\rowcolor{myblue}
&LLaDA2.0-Uni &D-Diff. + Diff. &0.54 &0.77 &0.82 &0.79 &0.87 &0.60 &0.68 \\
\rowcolor{myblue}
&~+ w/ thinking & D-Diff. + Diff. & 0.73  & 0.79 & 0.86 & 0.81 & 0.88 & 0.74 & 0.78 \\
\bottomrule
\end{tabularx}
\label{table:wise_benchmark}
\vspace{-0.06in}
\end{table}

\subsubsection{General Image Generation Performance}
\paragraph{GenEval.} Table~\ref{table:geneval} presents a comparison of model performance on the GenEval benchmark, which is designed to evaluate object-centric T2I generation using compositional prompts with diverse object attributes. 
LLaDA2.0-Uni demonstrates strong compositional capabilities, achieving an overall score of 0.89.
This performance is highly competitive, significantly outperforming all unified models and bridging the gap with top-tier generation-only models.
In particular, LLaDA2.0-Uni shows a clear advantage in spatial arrangement, securing the highest Position score (0.90) across all evaluated models.

\paragraph{DPG-Bench.} Table~\ref{table:dpg_benchmark} reports the text-to-image generation results on the DPG benchmark.
Notably, LLaDA2.0-Uni achieves the state-of-the-art overall score of 87.76 among unified models, outperforming strong baselines such as LLaDA-o (87.04) and HunyuanImage-3.0 (86.10).
Specifically, our model secures the highest scores on the Entity (93.55) and Other (94.04) sub-metrics.
Furthermore, LLaDA2.0-Uni delivers highly competitive performance even against specialized generation-only models, surpassing Z-Image-Turbo (84.86) and demonstrating robust text-image alignment capabilities.

\paragraph{One-IG Bench.} As shown in Table~\ref{table:oneig_en}, LLaDA2.0-Uni achieves a highly competitive overall score of 0.505 on OneIG-EN.
Notably, it outperforms all other unified models in Alignment (0.882) and Reasoning (0.323), reaching levels comparable to top dedicated generation models like Qwen-Image.
However, LLaDA2.0-Uni falls short of leading models in generating dense text, indicating an area for future improvement.

\paragraph{UniGenBench.} LLaDA2.0-Uni sets a new standard for unified models on UniGenBench (EN), achieving a top overall score of 79.63 (Table~\ref{table:unigen}).
It performs consistently well across all ten dimensions, showing a clear advantage in Logic (63.99) and Layout (90.30), where it even surpasses many specialized generation models.
These results demonstrate that LLaDA2.0-Uni effectively closes the performance gap between general-purpose unified models and top-tier specialized models.

\subsubsection{Text-Centric Image Generation Performance}
\paragraph{CVTG-2k.}
The CVTG-2K benchmark evaluates the capability of the model for text rendering across multiple regions.
As reported in Table~\ref{table:cvtg2k_rendering}, LLaDA2.0-Uni leads the unified models with an overall score of 0.765.
A key observation is its exceptional stability in multi-region text generation.
While baselines like BAGEL, Lumina-DiMOO, and InternVL-U show a sharp performance drop when the number of regions increases, our model experiences a much slower decline.

\subsubsection{Reasoning-Informed Image Generation Performance}
\paragraph{WISE-Bench.} 
The benefits of our large-scale multimodal pre-training are evident on the WISE benchmark (Table~\ref{table:wise_benchmark}), which evaluates complex semantic understanding and world knowledge in image generation.
LLaDA2.0-Uni achieves a strong overall score of 0.68, ranking first among all unified models and performing on par with generation-only models like LongCat-Image.
%
%
Notably, incorporating a reasoning mode yields an additional 10\% improvement. 
These results highlight the strong capability of LLaDA2.0-Uni in reasoning-informed generation.

\begin{table}[!tb]
\centering
\renewcommand{\arraystretch}{1.2}
\setlength\tabcolsep{1.5pt}
\footnotesize 
\newcolumntype{C}{>{\centering\arraybackslash}X}
\caption{Comparison of Instruction-based Image Editing Ability on ImgEdit Benchmark.}
\begin{tabularx}{\linewidth}{cc| *{9}{C} |c} 
\toprule
\textbf{Type} & \textbf{Model} & \textbf{Add} & \textbf{Adjust} &\textbf{Extract} &\textbf{Replace} &\textbf{Remove} &\textbf{Back.} &\textbf{Style} &\textbf{Hybrid} &\textbf{Action}
&\textbf{Overall $\uparrow$} \\
\midrule
\multirow{4}{*}{\rotatebox[origin=c]{90}{Gen. Only}} 
& FLUX.1 Kontext & 4.25 & 4.15 & 2.35 & 4.56 & 3.57 & 4.26 & 4.57 & 3.68 & 4.63 & 4.00 \\
& Step1X-Edit &3.88 &3.14 &1.76 &3.40 &2.41 &3.16 &4.63 &2.64 &2.52 &3.06 \\
& Qwen-Image-Edit &4.32 &4.36 &4.04 &4.64 &4.52 &4.37 &4.84 &3.39 &4.71 &4.35 \\
& Z-Image-Edit &4.40 &4.14 &4.30 &4.57 &4.13 &4.14 &4.85 &3.63 &4.50 &4.30 \\

\midrule
\multirow{5}{*}{\rotatebox[origin=c]{90}{Unified}} 
& BAGEL & 3.56 & 3.31 & 1.70 & 3.30 & 2.62 & 3.24 & 4.49 & 2.38 & 4.17 & 3.20 \\
& OmniGen2 & 3.57 & 3.06 & 1.77 & 3.74 & 3.20 & 3.57 & 4.81 & 2.52 & 4.68 & 3.44 \\
& InternVL-U &4.13 &3.40 &2.27 &4.13 &3.39 &3.84 &4.77 &3.03 &4.05 &3.67\\
& Lumina-DiMOO &3.41 &2.38 &1.90 &3.26 &2.21 &2.11 &4.19 &2.26 &3.17 &2.77\\
\rowcolor{myblue} 
& LLaDA2.0-Uni &3.76 &4.16 &2.40 & 4.04 & 3.82 & 4.07 & 4.60 & 3.97 & 4.42 & 3.92 \\
\bottomrule
\end{tabularx}
\label{table:imgedit_results}
\vspace{-0.15in}
\end{table}

\begin{table}[!tb]
\centering
\renewcommand{\arraystretch}{1.2}
\setlength\tabcolsep{2pt}
\small
\newcolumntype{C}{>{\centering\arraybackslash}X}
\vspace{-0.1in}
\caption{Comparison of Instruction-based Image Editing Ability on GEdit Benchmark. Abbreviations: Semantic Consistency (G\_SC), Perceptual Quality (G\_PQ), and Overall Score (G\_O).}
\begin{tabularx}{\linewidth}{ccc| *{3}{C} | *{3}{C}} 
\toprule
\multirow{2}{*}{\textbf{Type}} & \multirow{2}{*}{\textbf{Model}} & \multirow{2}{*}{\textbf{Arch.}} & \multicolumn{3}{c|}{\textbf{GEdit-Bench-EN$\uparrow$}} & \multicolumn{3}{c}{\textbf{GEdit-Bench-CN$\uparrow$}} \\
\cmidrule(lr){4-6} \cmidrule(lr){7-9}
& & & \textbf{G\_SC} & \textbf{G\_PQ} & \textbf{G\_O} & \textbf{G\_SC} & \textbf{G\_PQ} & \textbf{G\_O} \\
\midrule
\multirow{5}{*}{\rotatebox[origin=c]{90}{Gen. Only}} 
& FLUX.1 Kontext & Diff. & 6.52 & 7.38 &  6.00 & - & - & - \\
& Step1X-Edit & Diff. &7.66 &7.35 &6.97 &7.20 &6.87 &6.86 \\
& Qwen-Image-Edit & Diff. &8.00 & 7.86 & 7.56 & 7.82 & 7.79 & 7.52\\
&LongCat-Image-Edit & Diff. &8.18 &8.00 &7.64 &8.08 &7.99 &7.60 \\
&Z-Image-Edit & Diff. &8.11 &7.72 &7.57 &8.03 &7.80 &7.54 \\
\midrule
\multirow{5}{*}{\rotatebox[origin=c]{90}{Unified}} 
& BAGEL & AR + Diff. & 7.36 & 6.83 & 6.52 & 7.34 & 6.85 & 6.50 \\
& OmniGen2 & AR + Diff. & 7.16 & 6.77 & 6.41 & - & - & - \\
& InternVL-U & AR + Diff. & - & - & 6.66 & - & - & -\\
& Lumina-DiMOO  & D-Diff. &- &- &3.91 &- &- & -\\
\rowcolor{myblue} 
& LLaDA2.0-Uni & D-Diff. + Diff. &6.68 &7.52 &6.61 &6.63 &7.67 &6.66 \\
\bottomrule
\end{tabularx}
\label{table:gedit_bench}
\vspace{-0.08in}
\end{table}

\begin{table}[!tb]
\centering
\renewcommand{\arraystretch}{1.2}
\setlength\tabcolsep{3pt}
\small
\newcolumntype{C}{>{\centering\arraybackslash}X}

\caption{Comparison of Multi-Reference Image Editing Ability on MICo-Bench.}
\begin{tabularx}{\linewidth}{cc| *{4}{C}|c} 
\toprule

\textbf{Model} & \textbf{Arch.} & \textbf{Object} &\textbf{Person} &\textbf{HOI} &\textbf{De\&Re} &\textbf{~Overall$\uparrow$} \\

\midrule
Qwen-Image-Edit & Diff. & 52.4  & 21.1 & 35.0 & 37.4 & 35.9 \\

BAGEL & AR + Diff. & 39.0 & 28.5 & 25.3 & 44.5 & 34.4 \\

OmniGen2 & AR + Diff. & 46.3 & 22.9 & 32.2 & 36.8 & 33.8 \\

Lumina-DiMOO & D-Diff. & 38.4 & 12.1 & 24.7 & 21.3 & 23.3 \\
\rowcolor{myblue}

LLaDA2.0-Uni &D-Diff. + Diff. &51.0 &32.8 &46.0 &54.4 &47.1  \\
\bottomrule
\end{tabularx}
\label{table:mico-bench}
\vspace{-0.15in}
\end{table}

\subsection{Image Editing}
\subsubsection{Evaluation Settings}
\paragraph{Benchmarks.}
We evaluate LLaDA2.0-Uni on two general instruction-based image editing benchmarks: ImgEdit-Bench~\citep{ye2025imgedit} and GEdit-Bench~\citep{liu2025step1x-edit}. 
Additionally, we provide qualitative comparisons against leading models on MICo-Bench~\citep{wei2025mico}, a challenging multi-reference image editing benchmark.

\paragraph{Baselines.} 
We evaluate LLaDA2.0-Uni against specialized editing models (FLUX.1 Kontext~\citep{labs2025flux1kontextflowmatching}, Step1X-Edit~\citep{liu2025step1x-edit}, Qwen-Image-Edit~\citep{wu2025qwen}, Z-Image-Edit~\citep{cai2025z}) and unified models (BAGEL~\citep{bagel}, OmniGen2~\citep{wu2025omnigen2}, InternVL-U~\citep{tian2026internvl}, Lumina-DiMOO~\citep{dimoo}).
For the multi-reference editing benchmark, we compare with BAGEL, Qwen-Image-Edit, and OmniGen2, as they are the only baselines that natively support multiple image inputs.

\subsubsection{General Image Editing Performance}
\paragraph{ImgEdit.} Table~\ref{table:imgedit_results} presents the instruction-based image editing performance on the ImgEdit benchmark.
Among unified models, LLaDA2.0-Uni achieves the best Overall score of 3.92, ranking first and significantly outperforming peers like OmniGen2 (3.44) and InternVL-U (3.67).
Notably, LLaDA2.0-Uni excels in the Adjust and Hybrid tasks, securing the highest scores within the unified category. 
These results underscore its robust capability to comprehend and execute intricate editing instructions.

\paragraph{GEdit-Bench.} 
To further assess complex image editing capabilities, we evaluate our model on the GEdit benchmark.
As detailed in Table~\ref{table:gedit_bench}, LLaDA2.0-Uni achieves solid overall scores across both English (6.61) and Chinese (6.66) evaluations.
A key highlight is its strong performance in the Perceptual Quality category, proving that the model can execute edits without sacrificing the visual quality of the original image.

\subsubsection{Multi-Reference Image Editing Performance}
\paragraph{MICo-Bench.} 
To evaluate the multi-reference image composition capabilities of LLaDA2.0-Uni, we conduct experiments on the MICo-Bench.
As shown in Table~\ref{table:mico-bench}, LLaDA2.0-Uni achieves the best overall performance, setting a new state-of-the-art on this benchmark with a score of 47.1.
It significantly outperforms strong baselines such as OmniGen2 (33.8) and Qwen-Image (35.9).
Notably, Lumina-DiMOO, which shares the same dLLM architecture, struggles significantly on this task, yielding an overall score of only 23.3.
This stark contrast clearly validates the effectiveness of our architecture and data pipeline.

\subsection{Interleaved}
\subsubsection{Interleaved Generation}
\begin{figure}[tb]
    \centering
    \includegraphics[width=1.0\linewidth]{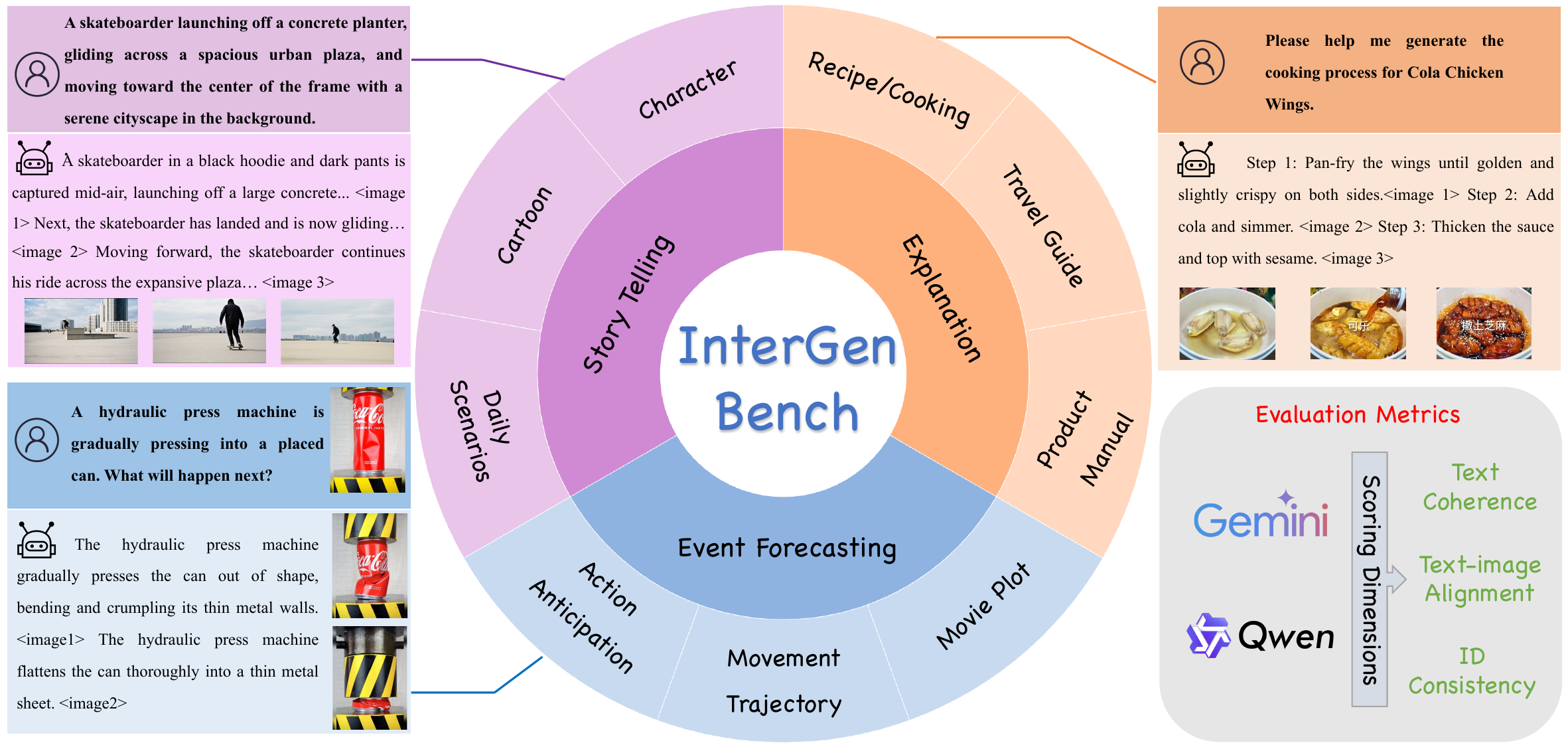}
    \caption{Overview of InterGen Benchmark.
    }
    \label{fig:intergen}
    \vspace{-0.1in}
\end{figure}

\paragraph{Benchmark Construction.} 
Existing models like Bagel and Nextflow support interleaved generation, yet a standard benchmark remains absent.
Datasets like ISG-BENCH~\citep{chen2024interleaved} and OpenING~\citep{zhou2025opening} contain interleaved cases but specialized tasks (e.g., 3D scene transformation) that most models cannot process.
To address this, we propose the InterGen benchmark.
As shown in Figure~\ref{fig:intergen}, it is structured into three main categories and various subcategories to comprehensively cover practical interleaved applications.
InterGen comprises 150 samples and utilizes advanced VLMs (Gemini-3 and Qwen3-VL) as a judge to assess performance across three dimensions: text coherence, text-image alignment, and ID consistency.

\begin{figure}[tb]
    \centering
    \includegraphics[width=1.0\linewidth]{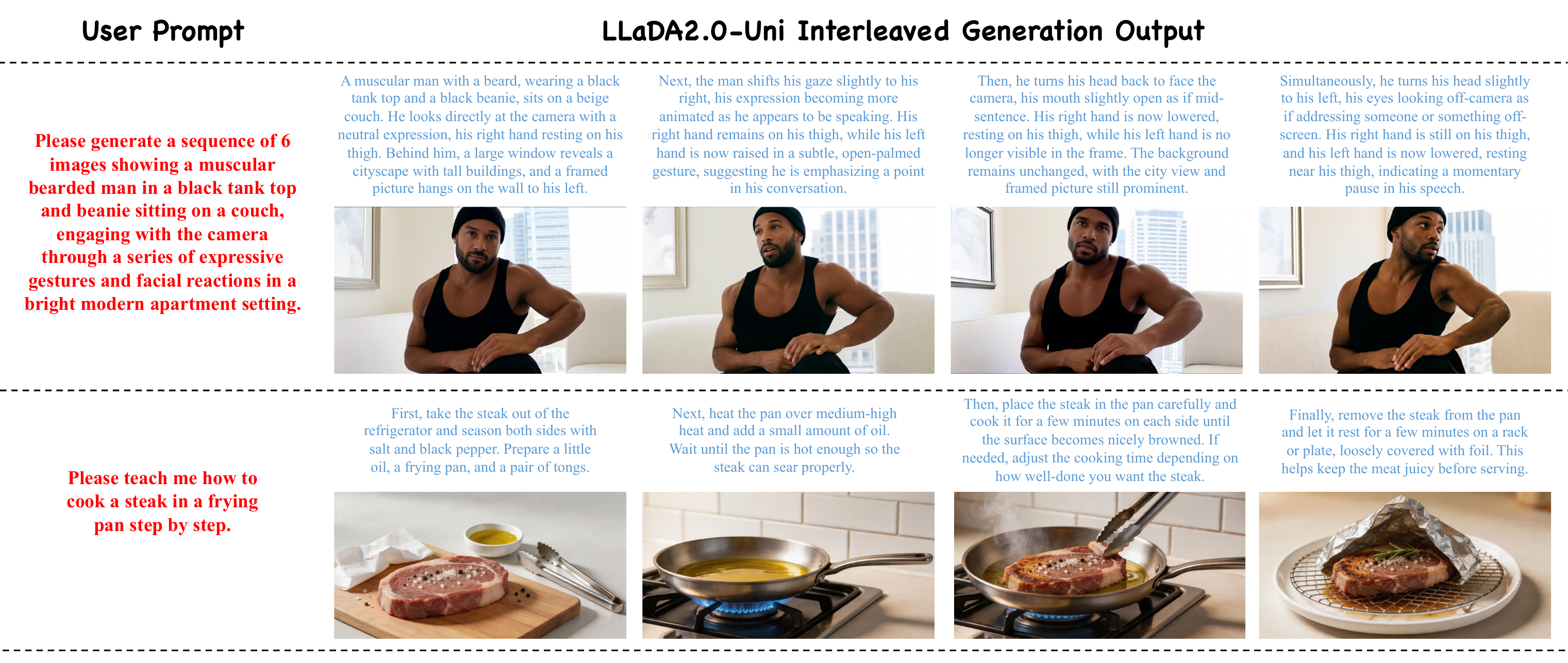}
    \vspace{-0.25in}
    \caption{Qualitative Results on Interleaved Generation Task.}
    \label{fig:inter_generation}
    \vspace{-0.1in}
\end{figure}

\begin{table}[t]
\centering
\renewcommand{\arraystretch}{1.2}
\setlength\tabcolsep{2pt}
\small
\newcolumntype{C}{>{\centering\arraybackslash}X}

\caption{Comparison of Interleaved Generation Ability on InterGen Benchmark.}
\begin{tabularx}{\linewidth}{cc| *{2}{C} | *{2}{C} |  *{2}{C}} 
\toprule
\multirow{2}{*}{\textbf{Model}} & \multirow{2}{*}{\textbf{Arch.}} & \multicolumn{2}{c|}{\textbf{Story Telling}} & \multicolumn{2}{c|}{\textbf{Explanation}} & \multicolumn{2}{c}{\textbf{Event Forecasting}} \\
& & \textbf{Gemini$\uparrow$} & \textbf{Qwen3-VL$\uparrow$} & \textbf{Gemini$\uparrow$} & \textbf{Qwen3-VL$\uparrow$} & \textbf{Gemini$\uparrow$} & \textbf{Qwen3-VL$\uparrow$}\\
\midrule

Emu3.5 & AR + Diff.  &6.28 & 6.83 &6.19 &6.48 &5.08  & 5.75  \\
\rowcolor{myblue}
LLaDA2.0-Uni & D-Diff. + Diff. 	&6.42 &7.02 &6.22 &6.35 &5.19	&5.94\\
\bottomrule
\end{tabularx}
\label{table:inter_bench}
\end{table}

\paragraph{InterGen Benchmark.} We primarily compare our model with Emu3.5~\citep{cui2025emu35nativemultimodalmodels}, as other models capable of interleaved generation (such as NextFlow~\citep{zhang2026nextflow} and Mogao~\citep{liao2025mogao}) are not yet open-source.
As shown in Table~\ref{table:inter_bench}, LLaDA2.0-Uni generally outperforms Emu3.5 on the InterGen benchmark.
Specifically, it achieves higher scores in the Story Telling and Time Series Forecasting tasks, while demonstrating comparable performance in Explanation.
More visualizations of the interleaved generation are shown in Figure~\ref{fig:inter_generation}.

\begin{figure}[!t]
    \centering
    \vspace{-0.1in}
    \includegraphics[width=1.0\linewidth]{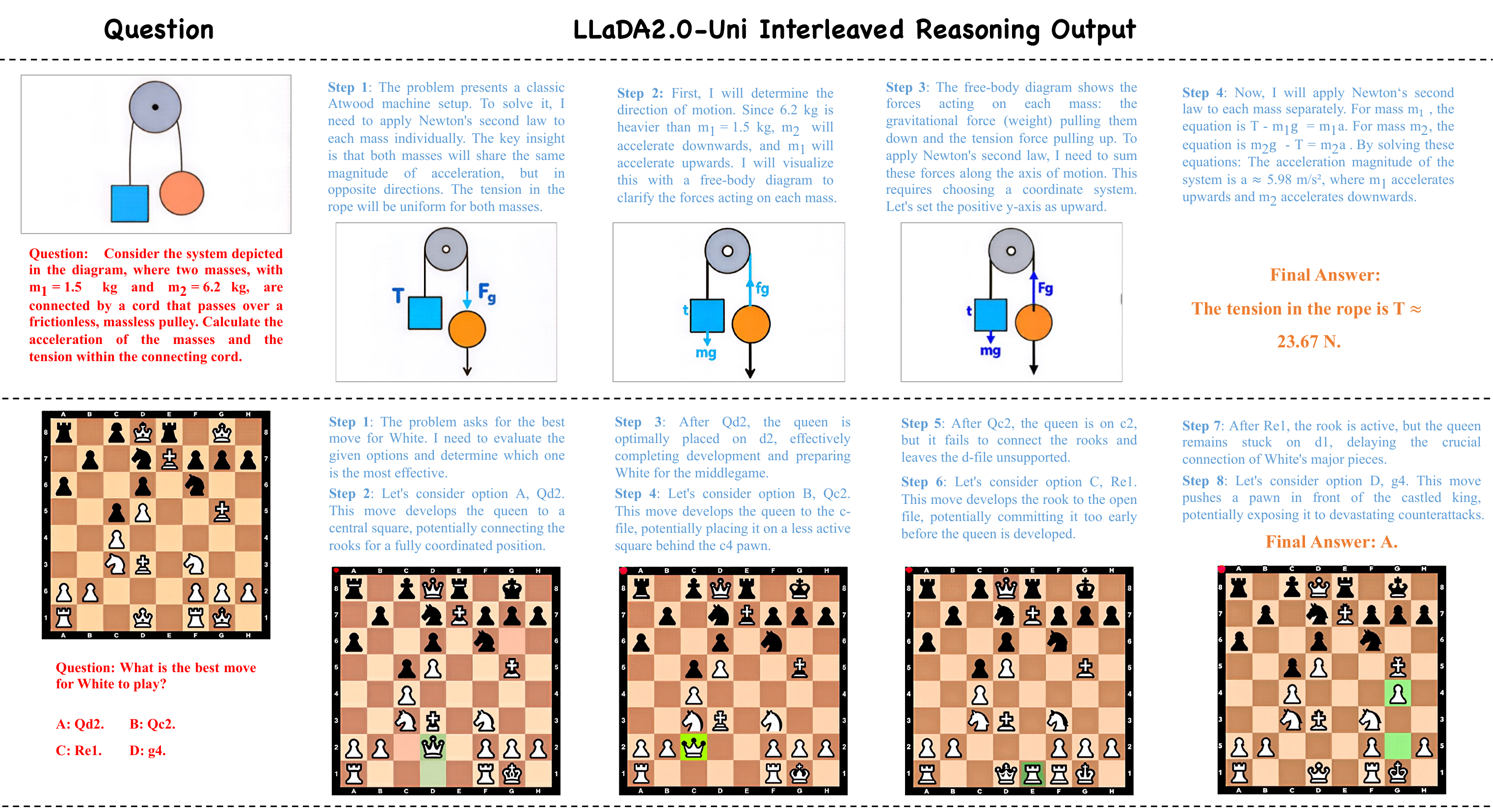}
    \vspace{-0.23in}
    \caption{Qualitative Results on Interleaved Reasoning Task.}
    \label{fig:inter_reason}
    \vspace{-0.13in}
\end{figure}

\subsubsection{Interleaved Reasoning}
Unified models have already achieved impressive results in standard image understanding and generation tasks.
Looking forward, exploring interleaved reasoning emerges as a critical bottleneck to overcome.
We believe this is an essential step toward building a strong synergy between visual generation and understanding.
In LLaDA 2.0-Uni, we conduct a preliminary exploration into interleaved reasoning capability.
As illustrated in Figure~\ref{fig:inter_reason}, through interleaved reasoning, our model successfully deduces logical strategies in in chess games and provides step-by-step solutions to physics problems.
These promising results give us great confidence to further expand interleaved reasoning capabilities in future research.

\subsection{Ablation Study}
\subsubsection{Analysis of SPRINT Acceleration} \label{sec:sprint_ablation}
\begin{table}[!b]
\centering
\vspace{-0.18in}
\renewcommand{\arraystretch}{1.22}
\caption{Performance and TPS comparison with and without SPRINT. (w/o SGLang)}
\label{tab:sprint}
\resizebox{\textwidth}{!}{%
\begin{tabular}{cccccccccccccc}
\toprule
\textbf{Method} & \textbf{Metric}
  & \textbf{AI2D} & \textbf{OCRB} & \textbf{MathVista}
  & \textbf{ChartQA} & \textbf{DocVQA} & \textbf{MMMU}
  & \textbf{MMStar} & \textbf{GenEval} & \textbf{DPG}
  & \textbf{Avg.} & \textbf{$\Delta$} \\
\midrule
\multirow{2}{*}{LLaDA2.0-Uni}
  & Score & 82.0 & 75.7 & 68.1 & 80.1 & 89.5 & 50.1 & 64.1 & 89.0 & 87.76 & 76.3 & -- \\
  & TPS   & 19.5 & 21.2 & 55.0 & 28.7 & 8.0  & 49.4 & 31.7 & 2.8  & 2.7  & 24.3 & -- \\
\midrule
\multirow{2}{*}{\ \ + SPRINT}
  & Score & 80.9 & 73.4 & 67.2 & 81.0 & 89.0 & 52.5 & 63.0 & 87.8 & 86.27 & 75.7 & $-$0.6 \\
  & TPS   & 42.9 & 36.0 & 75.0 & 62.3 & 27.6 & 52.2 & 49.2 & 5.1  & 7.8  & 39.8 & $\times$1.6 \\
\bottomrule
\end{tabular}%
}
\end{table}

Table~\ref{tab:sprint} evaluates SPRINT on nine multimodal benchmarks, covering both understanding (AI2D, OCRBench, MathVista, ChartQA, DocVQA, MMMU, MMStar) and generation (GenEval, DPG). 
SPRINT shifts the average score from 76.3 to 75.7 ($-$0.6) while accelerating generation from 24.3 to 39.8 TPS ($1.6\times$). 
Several benchmarks show score decreases; the most notable are OCRBench ($-2.3$) and DPG ($-1.5$).
OCRBench demands precise character-level prediction, where the lower threshold $\tau = 0.93$ may accept tokens before sufficient refinement. 
The speedup is largest on benchmarks with longer outputs, where the per-step savings from prefix pruning compound across many denoising iterations: DocVQA reaches $3.5\times$ (8.0 $\to$ 27.6), and ChartQA and AI2D both reach $2.2\times$.
SPRINT improves MMMU by +2.4 and ChartQA by +0.9. 
The non-uniform unmasking schedule concentrates refinement on uncertain positions, effectively increasing the denoising budget for difficult tokens without additional forward passes.
Furthermore, we are integrating LLaDA2.0-Uni with SGLang~\citep{Sglang} to accelerate inference, and the implementation will be made publicly available soon.

\subsubsection{Analysis of Diffusion Decoder} \label{sec:distilled_ablation}
Table~\ref{tab:distilled_perf} shows the performance and speed comparison between the Diffusion Decoder (50 steps) and Diffusion Decoder Turbo (8 steps) across five benchmarks. Speed is measured on a single GPU at $1024 \times 1024$ resolution with a batch size of 1, using BF16 precision.

\begin{table}[!htb]
\centering
\setlength\tabcolsep{10pt}
\renewcommand{\arraystretch}{1.0}
\caption{Performance and Speed Comparison between Diffusion Decoder and Diffusion Decoder Turbo.}
\resizebox{1.0\textwidth}{!}{%
\begin{tabular}{cc|ccccc}
\toprule
\textbf{Method} & \textbf{Speed (s / img)}
  & \textbf{GenEval} & \textbf{DPG} & \textbf{UniGenBench}
  & \textbf{OneIG-EN}
  & \textbf{WISE} 
  \\ 
\midrule
Diffusion Decoder (50 steps)
  & 32.95 & 0.89 & 87.76 & 79.63 & 0.505 & 0.68 \\
\midrule
Diffusion Decoder Turbo (8 steps)
  & 2.90 & 0.87 & 87.24 & 79.76 &0.500 & 0.68  \\
\bottomrule
\end{tabular}%
\label{tab:distilled_perf}
}
\end{table}

\begin{figure}[!t]
    \centering
    \vspace{-0.1in}
    \includegraphics[width=1.0\linewidth]{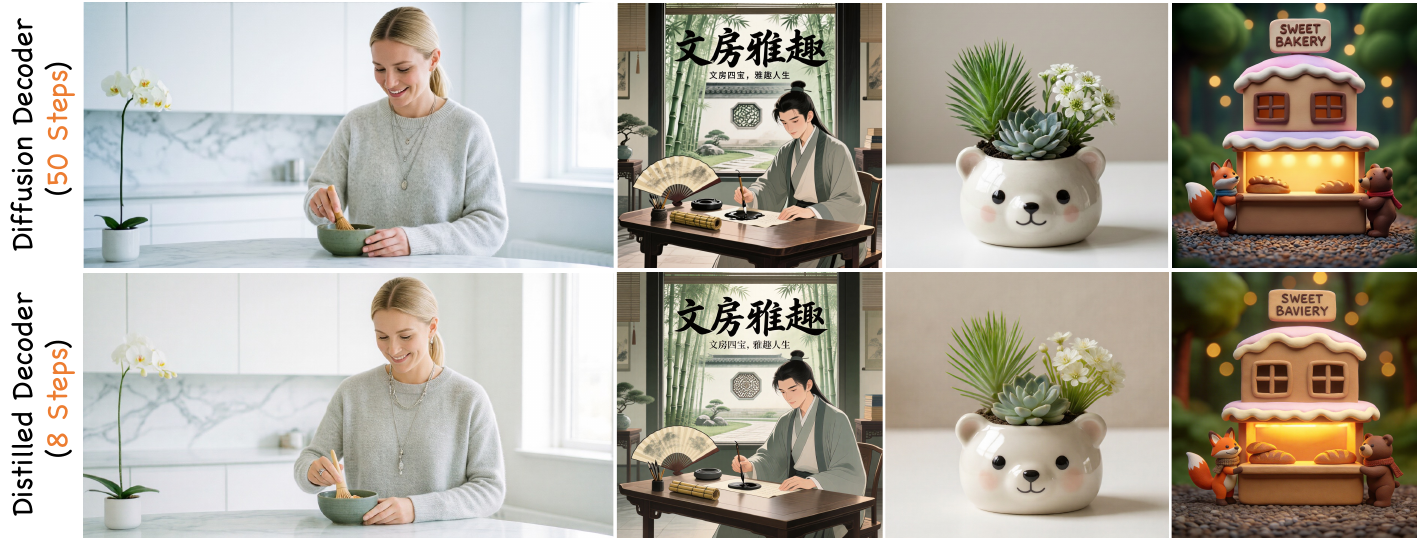}
    \vspace{-0.23in}
    \caption{Visual Comparison of the Decoder and the Distilled Version.}
    \label{fig:decoder_comp}
    \vspace{-0.13in}
\end{figure}

Through few-step acceleration, the Diffusion Decoder Turbo achieves an 11.4$\times$ speedup (from 32.95\,s/img to 2.90\,s/img) while maintaining competitive performance across all benchmarks.
Specifically, it retains a GenEval score of 0.87 (vs.\ 0.89 for the base) and a DPG score of 87.24 (vs.\ 87.76). 
Similarly, on the UniGenBench, OneIG-EN, and WISE benchmarks, Diffusion Decoder Turbo achieves performance on par with the original decoder.
Moreover, as shown in Figure~\ref{fig:decoder_comp}, the visual quality remains virtually indistinguishable.

\section{Conclusion and Future Directions}
In this work, we introduce LLaDA2.0-Uni, a unified framework that enables both multimodal understanding and generation within a single diffusion large language model.
Built on the LLaDA 2.0 backbone, our approach uses a SigLIP-VQ tokenizer to map visual inputs into semantically rich discrete tokens, allowing text and images to be modeled in a shared space.
Extensive experiments show that LLaDA2.0-Uni achieves strong performance across various benchmarks, including multimodal understanding, image generation, and editing.
Furthermore, the model naturally supports interleaved generation and chain-of-thought reasoning, demonstrating the flexibility and practicality of our unified architecture.

Despite these advancements, several areas remain for future improvement: 
\begin{itemize}[leftmargin=16pt]
    \item \textbf{Enhancing Visual Detail.} While the SigLIP-VQ tokenizer provides rich semantic information, it struggles to preserve fine-grained image details. Future work could focus on better reconstruction techniques to benefit detail-sensitive tasks like image editing.

    \item \textbf{Scaling Interleaved Capabilities.} To fully unlock the model’s potential for complex interleaved generation and reasoning, further scaling of training data and model capacity is required.

    \item \textbf{Reinforcement Learning.} Although we have begun exploring RL for unified dLLMs, optimizing its performance remains a challenge. We plan to further refine the RL framework in future versions and then release it to the community.   
\end{itemize}

\clearpage
\bibliographystyle{antgroup}
\bibliography{ref/reference}

@String(IJCV = {Int. J. Comput. Vis.})

@String(CVPR= {IEEE Conf. Comput. Vis. Pattern Recog.})

@String(ICCV= {Int. Conf. Comput. Vis.})

@String(ECCV= {Eur. Conf. Comput. Vis.})

@String(NIPS= {Adv. Neural Inform. Process. Syst.})

@String(ICLR = {Int. Conf. Learn. Represent.})

@String(AAAI = {AAAI})

@String{ijcv = "International Journal of Computer Vision (IJCV)"}

@String{cvpr = "Proceedings of the IEEE Conference on Computer Vision and Pattern Recognition (CVPR)"}

@String{nips = "Advances in Neural Information Processing Systems (NeurIPS)"}

@String{aaai = "Proceedings of the AAAI Conference on Artificial Intelligence (AAAI)"}

@String{eccv = "Proceedings of the European Conference on Computer Vision (ECCV)"}

@String{iclr = "Proceedings of the International Conference on Learning Representations (ICLR)"}

@String{iccv = "Proceedings of the IEEE International Conference on Computer Vision (ICCV)"}

@String{wacv = "Proceedings of the IEEE Winter Conference on Applications of Computer Vision (WACV)"}

@String{acl = "Proceedings of the Annual Meeting of the Association for Computational Linguistics (ACL)"}

@String{emnlp = "Proceedings of the Conference on Empirical Methods in Natural Language Processing (EMNLP)"}

@article{team2025qwen3VL,
  title={Qwen3-vl: Sharper vision, deeper thought, broader action},
  author={{Qwen Team}},
  journal={Qwen Blog. Accessed},
  pages={10--04},
  year={2025}
}

@article{LLaDA,
  title={Large language diffusion models},
  author={Nie, Shen and Zhu, Fengqi and You, Zebin and Zhang, Xiaolu and Ou, Jingyang and Hu, Jun and Zhou, Jun and Lin, Yankai and Wen, Ji-Rong and Li, Chongxuan},
  journal={arXiv preprint arXiv:2502.09992},
  year={2025}
}

@article{LLaDA2,
  title={Llada2. 0: Scaling up diffusion language models to 100b},
  author={Bie, Tiwei and Cao, Maosong and Chen, Kun and Du, Lun and Gong, Mingliang and Gong, Zhuochen and Gu, Yanmei and Hu, Jiaqi and Huang, Zenan and Lan, Zhenzhong and others},
  journal={arXiv preprint arXiv:2512.15745},
  year={2025}
}

@article{Ling2,
  title={Every activation boosted: Scaling general reasoner to 1 trillion open language foundation},
  author={{Ling Team}},
  journal={arXiv preprint arXiv:2510.22115},
  year={2025}
}

@article{bai2025qwen2,
  title={Qwen2. 5-vl technical report},
  author={{Qwen Team}},
  journal={arXiv preprint arXiv:2502.13923},
  year={2025}
}

@article{Paddleocr,
  title={Paddleocr 3.0 technical report},
  author={Cui, Cheng and Sun, Ting and Lin, Manhui and Gao, Tingquan and Zhang, Yubo and Liu, Jiaxuan and Wang, Xueqing and Zhang, Zelun and Zhou, Changda and Liu, Hongen and others},
  journal={arXiv preprint arXiv:2507.05595},
  year={2025}
}

@inproceedings{Objects365,
  title={Objects365: A large-scale, high-quality dataset for object detection},
  author={Shao, Shuai and Li, Zeming and Zhang, Tianyuan and Peng, Chao and Yu, Gang and Zhang, Xiangyu and Li, Jing and Sun, Jian},
  booktitle=iccv,
  year={2019}
}

@article{openimages,
  title={The open images dataset v4: Unified image classification, object detection, and visual relationship detection at scale},
  author={Kuznetsova, Alina and Rom, Hassan and Alldrin, Neil and Uijlings, Jasper and Krasin, Ivan and Pont-Tuset, Jordi and Kamali, Shahab and Popov, Stefan and Malloci, Matteo and Kolesnikov, Alexander and others},
  journal=ijcv,
  year={2020}
}

@inproceedings{kazemzadeh2014referitgame,
  title={Referitgame: Referring to objects in photographs of natural scenes},
  author={Kazemzadeh, Sahar and Ordonez, Vicente and Matten, Mark and Berg, Tamara},
  booktitle=emnlp,
  year={2014}
}

@inproceedings{yu2016modeling,
  title={Modeling context in referring expressions},
  author={Yu, Licheng and Poirson, Patrick and Yang, Shan and Berg, Alexander C and Berg, Tamara L},
  booktitle=eccv,
  year={2016},

}

@article{deepseekv3,
  title={Deepseek-v3 technical report},
  author={Liu, Aixin and Feng, Bei and Xue, Bing and Wang, Bingxuan and Wu, Bochao and Lu, Chengda and Zhao, Chenggang and Deng, Chengqi and Zhang, Chenyu and Ruan, Chong and others},
  journal={arXiv preprint arXiv:2412.19437},
  year={2024}
}

@article{sjl_moe,
  title={MoE Travels 3},
  author = {Jianlin Su},
  howpublished = {\url{https://kexue.fm/archives/10757}},
  year = {2025},
}

@article{li2025lavida,
  title={Lavida: A large diffusion language model for multimodal understanding},
  author={Li, Shufan and Kallidromitis, Konstantinos and Bansal, Hritik and Gokul, Akash and Kato, Yusuke and Kozuka, Kazuki and Kuen, Jason and Lin, Zhe and Chang, Kai-Wei and Grover, Aditya},
  journal={arXiv preprint arXiv:2505.16839},
  year={2025}
}

@article{dFactory,
  title={dFactory: Easy and Efficient dLLM Fine-Tuning},
  author = {InclusionAI},
  howpublished = {\url{https://github.com/inclusionAI/dFactory}},
  year = {2025},
}

@article{ma2025veomni,
  title={Veomni: Scaling any modality model training with model-centric distributed recipe zoo},
  author={Ma, Qianli and Zheng, Yaowei and Shi, Zhelun and Zhao, Zhongkai and Jia, Bin and Huang, Ziyue and Lin, Zhiqi and Li, Youjie and Yang, Jiacheng and Peng, Yanghua and others},
  journal={arXiv preprint arXiv:2508.02317},
  year={2025}
}

@article{you2025llada,
  title={Llada-v: Large language diffusion models with visual instruction tuning},
  author={You, Zebin and Nie, Shen and Zhang, Xiaolu and Hu, Jun and Zhou, Jun and Lu, Zhiwu and Wen, Ji-Rong and Li, Chongxuan},
  journal={arXiv preprint arXiv:2505.16933},
  year={2025}
}

@article{mathvista,
  title={Mathvista: Evaluating mathematical reasoning of foundation models in visual contexts},
  author={Lu, Pan and Bansal, Hritik and Xia, Tony and Liu, Jiacheng and Li, Chunyuan and Hajishirzi, Hannaneh and Cheng, Hao and Chang, Kai-Wei and Galley, Michel and Gao, Jianfeng},
  journal={arXiv preprint arXiv:2310.02255},
  year={2023}
}

@inproceedings{We-math,
  title={We-math: Does your large multimodal model achieve human-like mathematical reasoning?},
  author={Qiao, Runqi and Tan, Qiuna and Dong, Guanting and MinhuiWu, MinhuiWu and Sun, Chong and Song, Xiaoshuai and Wang, Jiapeng and Gongque, Zhuoma and Lei, Shanglin and Zhang, Yifan and others},
  booktitle=acl,
  year={2025}
}

@article{wang2024measuring,
  title={Measuring multimodal mathematical reasoning with math-vision dataset},
  author={Wang, Ke and Pan, Junting and Shi, Weikang and Lu, Zimu and Ren, Houxing and Zhou, Aojun and Zhan, Mingjie and Li, Hongsheng},
  journal=nips,
  year={2024}
}

@inproceedings{zhang2024mathverse,
  title={Mathverse: Does your multi-modal llm truly see the diagrams in visual math problems?},
  author={Zhang, Renrui and Jiang, Dongzhi and Zhang, Yichi and Lin, Haokun and Guo, Ziyu and Qiu, Pengshuo and Zhou, Aojun and Lu, Pan and Chang, Kai-Wei and Qiao, Yu and others},
  booktitle=eccv,
  year={2024},
}

@inproceedings{yue2024mmmu,
  title={Mmmu: A massive multi-discipline multimodal understanding and reasoning benchmark for expert agi},
  author={Yue, Xiang and Ni, Yuansheng and Zhang, Kai and Zheng, Tianyu and others},
  booktitle=cvpr,
  year={2024}
}

@inproceedings{yue2025mmmupro,
  title={Mmmu-pro: A more robust multi-discipline multimodal understanding benchmark},
  author={Yue, Xiang and Zheng, Tianyu and Ni, Yuansheng and Wang, Yubo and Zhang, Kai and Tong, Shengbang and Sun, Yuxuan and Yu, Botao and Zhang, Ge and Sun, Huan and others},
  booktitle=acl,
  year={2025}
}

@inproceedings{Simplevqa,
  title={Simplevqa: Multimodal factuality evaluation for multimodal large language models},
  author={Cheng, Xianfu and Zhang, Wei and Zhang, Shiwei and Yang, Jian and Guan, Xiangyuan and Wu, Xianjie and Li, Xiang and Zhang, Ge and Liu, Jiaheng and Mai, Yuying and others},
  booktitle=iccv,
  year={2025}
}

@inproceedings{Hallusionbench,
  title={Hallusionbench: an advanced diagnostic suite for entangled language hallucination and visual illusion in large vision-language models},
  author={Guan, Tianrui and Liu, Fuxiao and Wu, Xiyang and Xian, Ruiqi and Li, Zongxia and Liu, Xiaoyu and Wang, Xijun and Chen, Lichang and Huang, Furong and Yacoob, Yaser and others},
  booktitle=cvpr,
  year={2024}
}

@inproceedings{Mmbench,
  title={Mmbench: Is your multi-modal model an all-around player?},
  author={Liu, Yuan and Duan, Haodong and Zhang, Yuanhan and Li, Bo and Zhang, Songyang and Zhao, Wangbo and Yuan, Yike and Wang, Jiaqi and He, Conghui and Liu, Ziwei and others},
  booktitle=eccv,
  year={2024}
}

@article{mmstar,
  title={Are we on the right way for evaluating large vision-language models?},
  author={Chen, Lin and Li, Jinsong and Dong, Xiaoyi and Zhang, Pan and Zang, Yuhang and Chen, Zehui and Duan, Haodong and Wang, Jiaqi and Qiao, Yu and Lin, Dahua and others},
  journal=nips,
  year={2024}
}

@article{RealworldQA,
  title={Grok 1.5v: A New Era in AI Understanding},
  author = {OpenAI},
  howpublished = {\url{https://x.ai/blog/grok-1.5v}},
  year = {2024},
}

@inproceedings{Chartqa,
  title={Chartqa: A benchmark for question answering about charts with visual and logical reasoning},
  author={Masry, Ahmed and Do, Xuan Long and Tan, Jia Qing and Joty, Shafiq and Hoque, Enamul},
  booktitle=acl,
  year={2022}
}

@inproceedings{Docvqa,
  title={Docvqa: A dataset for vqa on document images},
  author={Mathew, Minesh and Karatzas, Dimosthenis and Jawahar, CV},
  booktitle=wacv,
  year={2021}
}

@inproceedings{InfoVQA,
  title={Infographicvqa},
  author={Mathew, Minesh and Bagal, Viraj and Tito, Rub{\`e}n and Karatzas, Dimosthenis and Valveny, Ernest and Jawahar, CV},
  booktitle=wacv,
  year={2022}
}

@article{Charxiv,
  title={Charxiv: Charting gaps in realistic chart understanding in multimodal llms},
  author={Wang, Zirui and Xia, Mengzhou and He, Luxi and Chen, Howard and Liu, Yitao and Zhu, Richard and Liang, Kaiqu and Wu, Xindi and Liu, Haotian and Malladi, Sadhika and others},
  journal=nips,
  year={2024}
}

@article{Ocrbench,
  title={Ocrbench: on the hidden mystery of ocr in large multimodal models},
  author={Liu, Yuliang and Li, Zhang and Huang, Mingxin and Yang, Biao and Yu, Wenwen and Li, Chunyuan and Yin, Xu-Cheng and Liu, Cheng-Lin and Jin, Lianwen and Bai, Xiang},
  journal={Science China Information Sciences (SCIS)},
  year={2024}
}

@inproceedings{AI2D,
  title={A diagram is worth a dozen images},
  author={Kembhavi, Aniruddha and Salvato, Mike and Kolve, Eric and Seo, Minjoon and Hajishirzi, Hannaneh and Farhadi, Ali},
  booktitle=eccv,
  year={2016}
}

@inproceedings{countbench,
  title={Teaching clip to count to ten},
  author={Paiss, Roni and Ephrat, Ariel and Tov, Omer and Zada, Shiran and Mosseri, Inbar and Irani, Michal and Dekel, Tali},
  booktitle=iccv,
  year={2023}
}

@inproceedings{VLRewardBench,
  title={VL-RewardBench: A Challenging Benchmark for Vision-Language Generative Reward Models},
  author={Li, Lei and Wei, Yuancheng and Xie, Zhihui and Yang, Xuqing and Song, Yifan and Wang, Peiyi and An, Chenxin and Liu, Tianyu and Li, Sujian and Lin, Bill Yuchen and others},
  booktitle=cvpr,
  year={2025}
}

@article{Sglang,
  title={Sglang: Efficient execution of structured language model programs},
  author={Zheng, Lianmin and Yin, Liangsheng and Xie, Zhiqiang and Sun, Chuyue Livia and Huang, Jeff and Yu, Cody Hao and Cao, Shiyi and Kozyrakis, Christos and Stoica, Ion and Gonzalez, Joseph E and others},
  journal=nips,
  year={2024}
}

@article{agarwal2025gpt,
  title={gpt-oss-120b \& gpt-oss-20b model card},
  author={Agarwal, Sandhini and Ahmad, Lama and Ai, Jason and Altman, Sam and Applebaum, Andy and Arbus, Edwin and Arora, Rahul K and Bai, Yu and Baker, Bowen and Bao, Haiming and others},
  journal={arXiv preprint arXiv:2508.10925},
  year={2025}
}

@inproceedings{Internvl,
  title={Internvl: Scaling up vision foundation models and aligning for generic visual-linguistic tasks},
  author={Chen, Zhe and Wu, Jiannan and Wang, Wenhai and Su, Weijie and Chen, Guo and Xing, Sen and Zhong, Muyan and Zhang, Qinglong and Zhu, Xizhou and Lu, Lewei and others},
  booktitle=cvpr,
  year={2024}
}

@article{internvl35,
  title={Internvl3. 5: Advancing open-source multimodal models in versatility, reasoning, and efficiency},
  author={Wang, Weiyun and Gao, Zhangwei and Gu, Lixin and Pu, Hengjun and Cui, Long and Wei, Xingguang and Liu, Zhaoyang and Jing, Linglin and Ye, Shenglong and Shao, Jie and others},
  journal={arXiv preprint arXiv:2508.18265},
  year={2025}
}

@article{internvl2,
  title={Expanding performance boundaries of open-source multimodal models with model, data, and test-time scaling},
  author={Chen, Zhe and Wang, Weiyun and Cao, Yue and Liu, Yangzhou and Gao, Zhangwei and Cui, Erfei and Zhu, Jinguo and Ye, Shenglong and Tian, Hao and Liu, Zhaoyang and others},
  journal={arXiv preprint arXiv:2412.05271},
  year={2024}
}

@article{flux,
author = {{Black-forest-labs}},
  title = {{FLUX.1}},
  howpublished = {\url{https://bfl.ai/models/flux-kontext}},
  year = {2024},
}

@article{bagel,
  title={Emerging properties in unified multimodal pretraining},
  author={Deng, Chaorui and Zhu, Deyao and Li, Kunchang and Gou, Chenhui and Li, Feng and Wang, Zeyu and Zhong, Shu and Yu, Weihao and Nie, Xiaonan and Song, Ziang and others},
  journal={arXiv preprint arXiv:2505.14683},
  year={2025}
}

@article{dimoo,
  title={Lumina-dimoo: An omni diffusion large language model for multi-modal generation and understanding},
  author={Xin, Yi and Qin, Qi and Luo, Siqi and Zhu, Kaiwen and Yan, Juncheng and Tai, Yan and Lei, Jiayi and Cao, Yuewen and Wang, Keqi and Wang, Yibin and others},
  journal={arXiv preprint arXiv:2510.06308},
  year={2025}
}

@article{Mmada,
  title={Mmada: Multimodal large diffusion language models},
  author={Yang, Ling and Tian, Ye and Li, Bowen and Zhang, Xinchen and Shen, Ke and Tong, Yunhai and Wang, Mengdi},
  journal={arXiv preprint arXiv:2505.15809},
  year={2025}
}

@article{xomni,
  title={X-omni: Reinforcement learning makes discrete autoregressive image generative models great again},
  author={Geng, Zigang and Wang, Yibing and Ma, Yeyao and Li, Chen and Rao, Yongming and Gu, Shuyang and Zhong, Zhao and Lu, Qinglin and Hu, Han and Zhang, Xiaosong and others},
  journal={arXiv preprint arXiv:2507.22058},
  year={2025}
}

@article{lou2023discrete,
  title={Discrete diffusion modeling by estimating the ratios of the data distribution},
  author={Lou, Aaron and Meng, Chenlin and Ermon, Stefano},
  journal={arXiv preprint arXiv:2310.16834},
  year={2023}
}

@article{sahoo2024simple,
  title={Simple and effective masked diffusion language models},
  author={Sahoo, Subham and Arriola, Marianne and Schiff, Yair and Gokaslan, Aaron and Marroquin, Edgar and Chiu, Justin and Rush, Alexander and Kuleshov, Volodymyr},
  journal=nips,
  year={2024}
}

@inproceedings{Janus,
  title={Janus: Decoupling visual encoding for unified multimodal understanding and generation},
  author={Wu, Chengyue and Chen, Xiaokang and Wu, Zhiyu and Ma, Yiyang and Liu, Xingchao and Pan, Zizheng and Liu, Wen and others},
  booktitle=cvpr,
  year={2025}
}

@article{cui2025emu3,
  title={Emu3. 5: Native multimodal models are world learners},
  author={Cui, Yufeng and Chen, Honghao and Deng, Haoge and Huang, Xu and Li, Xinghang and Liu, Jirong and Liu, Yang and Luo, Zhuoyan and Wang, Jinsheng and Wang, Wenxuan and others},
  journal={arXiv preprint arXiv:2510.26583},
  year={2025}
}

@article{gu2025ui,
  title={Ui-venus technical report: Building high-performance ui agents with rft},
  author={Gu, Zhangxuan and Zeng, Zhengwen and Xu, Zhenyu and Zhou, Xingran and Shen, Shuheng and Liu, Yunfei and Zhou, Beitong and Meng, Changhua and Xia, Tianyu and Chen, Weizhi and others},
  journal={arXiv preprint arXiv:2508.10833},
  year={2025}
}

@article{wu2025qwen,
  title={Qwen-image technical report},
  author={Wu, Chenfei and Li, Jiahao and Zhou, Jingren and Lin, Junyang and Gao, Kaiyuan and Yan, Kun and Yin, Sheng-ming and Bai, Shuai and Xu, Xiao and Chen, Yilei and others},
  journal={arXiv preprint arXiv:2508.02324},
  year={2025}
}

@article{ming_flash,
  title={Ming-flash-omni: A sparse, unified architecture for multimodal perception and generation},
  author={AI, Inclusion and Ma, Bowen and Zou, Cheng and Yan, Canxiang and Jin, Chunxiang and Shen, Chunjie and Lian, Chenyu and Zheng, Dandan and Wang, Fudong and Xu, Furong and others},
  journal={arXiv preprint arXiv:2510.24821},
  year={2025}
}

@article{tschannen2025siglip,
  title={Siglip 2: Multilingual vision-language encoders with improved semantic understanding, localization, and dense features},
  author={Tschannen, Michael and Gritsenko, Alexey and Wang, Xiao and Naeem, Muhammad Ferjad and Alabdulmohsin, Ibrahim and Parthasarathy, Nikhil and Evans, Talfan and Beyer, Lucas and Xia, Ye and Mustafa, Basil and others},
  journal={arXiv preprint arXiv:2502.14786},
  year={2025}
}

@inproceedings{esser2021taming,
  title={Taming transformers for high-resolution image synthesis},
  author={Esser, Patrick and Rombach, Robin and Ommer, Bjorn},
  booktitle=cvpr,
  year={2021}
}

@article{wang2024emu3,
  title={Emu3: Next-token prediction is all you need},
  author={Wang, Xinlong and Zhang, Xiaosong and Luo, Zhengxiong and Sun, Quan and Cui, Yufeng and Wang, Jinsheng and Zhang, Fan and Wang, Yueze and Li, Zhen and Yu, Qiying and others},
  journal={arXiv preprint arXiv:2409.18869},
  year={2024}
}

@article{you2026lladao,
  title={LLaDA-o: An Effective and Length-Adaptive Omni Diffusion Model},
  author={You, Zebin and Zhang, Xiaolu and Zhou, Jun and Li, Chongxuan and Wen, Ji-Rong},
  journal={arXiv preprint arXiv:2603.01068},
  year={2026}
}

@article{tian2026internvl,
  title={InternVL-U: Democratizing Unified Multimodal Models for Understanding, Reasoning, Generation and Editing},
  author={Tian, Changyao and Yang, Danni and Chen, Guanzhou and Cui, Erfei and Wang, Zhaokai and Duan, Yuchen and Yin, Penghao and Chen, Sitao and Yang, Ganlin and Liu, Mingxin and others},
  journal={arXiv preprint arXiv:2603.09877},
  year={2026}
}

@inproceedings{qin2025lumina,
  title={Lumina-image 2.0: A unified and efficient image generative framework},
  author={Qin, Qi and Zhuo, Le and Xin, Yi and Du, Ruoyi and Li, Zhen and Fu, Bin and Lu, Yiting and Li, Xinyue and Liu, Dongyang and Zhu, Xiangyang and others},
  booktitle=iccv,
  year={2025}
}

@article{team2025longcat,
  title={Longcat-image technical report},
  author={Team, Meituan LongCat and Ma, Hanghang and Tan, Haoxian and Huang, Jiale and Wu, Junqiang and He, Jun-Yan and Gao, Lishuai and Xiao, Songlin and Wei, Xiaoming and Ma, Xiaoqi and others},
  journal={arXiv preprint arXiv:2512.07584},
  year={2025}
}

@article{cai2025z,
  title={Z-image: An efficient image generation foundation model with single-stream diffusion transformer},
  author={Cai, Huanqia and Cao, Sihan and Du, Ruoyi and Gao, Peng and Hoi, Steven and Hou, Zhaohui and Huang, Shijie and Jiang, Dengyang and Jin, Xin and Li, Liangchen and others},
  journal={arXiv preprint arXiv:2511.22699},
  year={2025}
}

@article{gao2025seedream,
  title={Seedream 3.0 technical report},
  author={Gao, Yu and Gong, Lixue and Guo, Qiushan and Hou, Xiaoxia and Lai, Zhichao and Li, Fanshi and Li, Liang and Lian, Xiaochen and Liao, Chao and Liu, Liyang and others},
  journal={arXiv preprint arXiv:2504.11346},
  year={2025}
}

@article{xin2025lumina,
  title={Lumina-mgpt 2.0: Stand-alone autoregressive image modeling},
  author={Xin, Yi and Yan, Juncheng and Qin, Qi and Li, Zhen and Liu, Dongyang and Li, Shicheng and Huang, Victor Shea-Jay and Zhou, Yupeng and Zhang, Renrui and Zhuo, Le and others},
  journal={arXiv preprint arXiv:2507.17801},
  year={2025}
}

@article{wu2025omnigen2,
  title={Omnigen2: Exploration to advanced multimodal generation},
  author={Wu, Chenyuan and Zheng, Pengfei and Yan, Ruiran and Xiao, Shitao and Luo, Xin and Wang, Yueze and Li, Wanli and Jiang, Xiyan and Liu, Yexin and Zhou, Junjie and others},
  journal={arXiv preprint arXiv:2506.18871},
  year={2025}
}

@article{cao2025hunyuanimage,
  title={Hunyuanimage 3.0 technical report},
  author={Cao, Siyu and Chen, Hangting and Chen, Peng and Cheng, Yiji and Cui, Yutao and Deng, Xinchi and Dong, Ying and Gong, Kipper and Gu, Tianpeng and Gu, Xiusen and others},
  journal={arXiv preprint arXiv:2509.23951},
  year={2025}
}

@article{zhang2026nextflow,
  title={NextFlow: Unified Sequential Modeling Activates Multimodal Understanding and Generation},
  author={Zhang, Huichao and Qu, Liao and Liu, Yiheng and Chen, Hang and Song, Yangyang and Dong, Yongsheng and Sun, Shikun and Li, Xian and Wang, Xu and Jiang, Yi and others},
  journal={arXiv preprint arXiv:2601.02204},
  year={2026}
}

@article{ghosh2023geneval,
  title={Geneval: An object-focused framework for evaluating text-to-image alignment},
  author={Ghosh, Dhruba and Hajishirzi, Hannaneh and Schmidt, Ludwig},
  journal=nips,
  year={2023}
}

@article{chang2025oneig,
  title={OneIG-Bench: Omni-dimensional Nuanced Evaluation for Image Generation}, 
  author={Jingjing Chang and Yixiao Fang and Peng Xing and Shuhan Wu and Wei Cheng and Rui Wang and Xianfang Zeng and Gang Yu and Hai-Bao Chen},
  journal={arXiv preprint arxiv:2506.07977},
  year={2025}
}

@article{UniGenBench++,
  title={UniGenBench++: A Unified Semantic Evaluation Benchmark for Text-to-Image Generation},
  author={Wang, Yibin and Li, Zhimin and Zang, Yuhang and Bu, Jiazi and Zhou, Yujie and Xin, Yi and He, Junjun and Wang, Chunyu and Lu, Qinglin and Jin, Cheng and others},
  journal={arXiv preprint arXiv:2510.18701},
  year={2025}
}

@article{du2025textcrafter,
  title={TextCrafter: Accurately Rendering Multiple Texts in Complex Visual Scenes},
  author={Du, Nikai and Chen, Zhennan and Chen, Zhizhou and Gao, Shan and Chen, Xi and Jiang, Zhengkai and Yang, Jian and Tai, Ying},
  journal={arXiv preprint arXiv:2503.23461},
  year={2025}
}

@article{niu2025wise,
  title={WISE: A World Knowledge-Informed Semantic Evaluation for Text-to-Image Generation},
  author={Niu, Yuwei and Ning, Munan and Zheng, Mengren and Jin, Weiyang and Lin, Bin and Jin, Peng and Liao, Jiaqi and Ning, Kunpeng and Feng, Chaoran and Zhu, Bin and Yuan, Li},
  journal={arXiv preprint arXiv:2503.07265},
  year={2025}
}

@article{hu2024ella,
  title={ELLA: Equip Diffusion Models with LLM for Enhanced Semantic Alignment}, 
  author={Xiwei Hu and Rui Wang and Yixiao Fang and Bin Fu and Pei Cheng and Gang Yu},
  year={2024},
  journal={arXiv preprint arXiv:2403.05135},
}

@article{ye2025imgedit,
  title={Imgedit: A unified image editing dataset and benchmark},
  author={Ye, Yang and He, Xianyi and Li, Zongjian and Lin, Bin and Yuan, Shenghai and Yan, Zhiyuan and Hou, Bohan and Yuan, Li},
  journal={arXiv preprint arXiv:2505.20275},
  year={2025}
}

@article{liu2025step1x-edit,
  title={Step1X-Edit: A Practical Framework for General Image Editing}, 
  author={Shiyu Liu and Yucheng Han and Peng Xing and Fukun Yin and Rui Wang and Wei Cheng and Jiaqi Liao and Yingming Wang and Honghao Fu and Chunrui Han and Guopeng Li and Yuang Peng and Quan Sun and Jingwei Wu and Yan Cai and Zheng Ge and Ranchen Ming and Lei Xia and Xianfang Zeng and Yibo Zhu and Binxing Jiao and Xiangyu Zhang and Gang Yu and Daxin Jiang},
  journal={arXiv preprint arXiv:2504.17761},
  year={2025}
}

@article{wei2025mico,
  title={MICo-150K: A Comprehensive Dataset Advancing Multi-Image Composition},
  author={Wei, Xinyu and Cen, Kangrui and Wei, Hongyang and Guo, Zhen and Li, Bairui and Wang, Zeqing and Zhang, Jinrui and Zhang, Lei},
  journal={arXiv preprint arXiv:2512.07348},
  year={2025}
}

@article{labs2025flux1kontextflowmatching,
  title={FLUX.1 Kontext: Flow Matching for In-Context Image Generation and Editing in Latent Space},
  author={Black Forest Labs and Stephen Batifol and Andreas Blattmann and Frederic Boesel and Saksham Consul and Cyril Diagne and Tim Dockhorn and Jack English and Zion English and Patrick Esser and Sumith Kulal and Kyle Lacey and Yam Levi and Cheng Li and Dominik Lorenz and Jonas Müller and Dustin Podell and Robin Rombach and Harry Saini and Axel Sauer and Luke Smith},
  journal={arXiv preprint arXiv:2506.15742},
  year={2025}
}

@article{fu2023mme,
  title={MME: A Comprehensive Evaluation Benchmark for Multimodal Large Language Models},
  author={Fu, Chaoyou and Chen, Peixian and Shen, Yunhang and Qin, Yulei and Zhang, Mengdan and Lin, Xu and Yang, Jinrui and Zheng, Xiawu and Li, Ke and Sun, Xing and others},
  journal={arXiv preprint arXiv:2306.13394},
  year={2023}
}

@article{vstar,
  title={V*: Guided Visual Search as a Core Mechanism in Multimodal LLMs},
  author={Penghao Wu and Saining Xie},
  journal={arXiv preprint arXiv:2312.14135},
  year={2023}
}

@inproceedings{wang2025koala,
  title={Koala-36m: A large-scale video dataset improving consistency between fine-grained conditions and video content},
  author={Wang, Qiuheng and Shi, Yukai and Ou, Jiarong and Chen, Rui and Lin, Ke and Wang, Jiahao and Jiang, Boyuan and Yang, Haotian and Zheng, Mingwu and Tao, Xin and others},
  booktitle=cvpr,
  year={2025}
}

@article{sun2026duality,
  title={Duality Models: An Embarrassingly Simple One-step Generation Paradigm},
  author={Sun, Peng and Shang, Xinyi and Lin, Tao and Shen, Zhiqiang},
  journal={arXiv preprint arXiv:2602.17682},
  year={2026}
}

@article{arriola2025block,
    title={Block Diffusion: Interpolating Between Autoregressive and Diffusion Language Models},
    author={Marianne Arriola and Aaron Gokaslan and Justin T Chiu and Zhihan Yang and Zhixuan Qi and Jiaqi Han and Subham Sekhar Sahoo and Volodymyr Kuleshov},
    journal=iclr,
    year={2025}
}

@article{su2021roformer,
      title={RoFormer: Enhanced Transformer with Rotary Position Embedding}, 
      author={Jianlin Su and Yu Lu and Shengfeng Pan and Bo Wen and Yunfeng Liu},
      journal={arXiv preprint arXiv:2104.09864},
      year={2021},
}

@article{cao2025artimusefinegrainedimageaesthetics,
      title={ArtiMuse: Fine-Grained Image Aesthetics Assessment with Joint Scoring and Expert-Level Understanding}, 
      author={Shuo Cao and Nan Ma and Jiayang Li and Xiaohui Li and Lihao Shao and Kaiwen Zhu and Yu Zhou and Yuandong Pu and Jiarui Wu and Jiaquan Wang and Bo Qu and Wenhai Wang and Yu Qiao and Dajuin Yao and Yihao Liu},
      journal={arXiv preprint arXiv:2507.14533},
      year={2025}
}

@inproceedings{deqa_score,
    title={Teaching Large Language Models to Regress Accurate Image Quality Scores using Score Distribution},
    author={You, Zhiyuan and Cai, Xin and Gu, Jinjin and Xue, Tianfan and Dong, Chao},
    booktitle=cvpr,
    year={2025}
}

@article{fang2025flux,
      title={FLUX-Reason-6M \& PRISM-Bench: A Million-Scale Text-to-Image Reasoning Dataset and Comprehensive Benchmark}, 
      author={Fang, Rongyao and Yu, Aldrich and Duan, Chengqi and Huang, Linjiang and Bai, Shuai and Cai, Yuxuan and Wang, Kun and Liu, Si and Liu, Xihui and Li, Hongsheng},
      journal={arXiv preprint arXiv:2509.09680},
      year={2025}
}

@inproceedings{li2026zebracot,
  title={Zebra-CoT: A Dataset for Interleaved Vision-Language Reasoning},
  author={Ang Li and Charles Wang and Deqing Fu and Kaiyu Yue and Zikui Cai
          and Wang Bill Zhu and Ollie Liu and Peng Guo and Willie Neiswanger
          and Furong Huang and Tom Goldstein and Micah Goldblum},
  booktitle=iclr,
  year={2026},
}

@article{weave2024,
  title={Weave: A Benchmark for Evaluating Multimodal Editing Models},
  author={Chow, Wei and Pan, Jiachun and Liang, Yongyuan and Zhou, Mingze and Song, Xue and Jia, Liyu and Zhang, Saining and Tang, Siliang and Li, Juncheng and Zhang, Fengda and others},
  journal={arXiv preprint arXiv:2511.15738},
  year={2024}
}

@inproceedings{LLaVA-OneVision-1.5,
  title={LLaVA-OneVision-1.5: Fully Open Framework for Democratized Multimodal Training},
  author={An, Xiang and Xie, Yin and Yang, Kaicheng and Zhang, Wenkang and Zhao, Xiuwei and Cheng, Zheng and Wang, Yirui and Xu, Songcen and Chen, Changrui and Wu, Chunsheng and Tan, Huajie and Li, Chunyuan and Yang, Jing and Yu, Jie and Wang, Xiyao and Qin, Bin and Wang, Yumeng and Yan, Zizhen and Feng, Ziyong and Liu, Ziwei and Li, Bo and Deng, Jiankang},
  booktitle={arXiv},  
  year={2025}
 }

@article{Penguin-VL,
  title={Penguin-VL: Exploring the Efficiency Limits of VLM with LLM-based Vision Encoders},
  author={Boqiang Zhang and Lei Ke and Ruihan Yang and Qi Gao and Tianyuan Qu and Rossell Chen and Dong Yu and Leoweiliang},
  journal={arXiv preprint arXiv:2603.06569},
  year={2026}
}

@inproceedings{ma2026x2edit,
  title={X2edit: Revisiting arbitrary-instruction image editing through self-constructed data and task-aware representation learning},
  author={Ma, Jian and Zhu, Xujie and Pan, Zihao and Peng, Qirong and Guo, Xu and Chen, Chen and Lu, Haonan},
  booktitle=aaai,
  year={2026}
}

@inproceedings{wei2024omniedit,
  title={Omniedit: Building image editing generalist models through specialist supervision},
  author={Wei, Cong and Xiong, Zheyang and Ren, Weiming and Du, Xeron and Zhang, Ge and Chen, Wenhu},
  booktitle=iclr,
  year={2024}
}

@article{lin2025uniworld,
  title={Uniworld-v1: High-resolution semantic encoders for unified visual understanding and generation},
  author={Lin, Bin and Li, Zongjian and Cheng, Xinhua and Niu, Yuwei and Ye, Yang and He, Xianyi and Yuan, Shenghai and Yu, Wangbo and Wang, Shaodong and Ge, Yunyang and others},
  journal={arXiv preprint arXiv:2506.03147},
  year={2025}
}

@article{zhuo2025factuality,
  title={Factuality Matters: When Image Generation and Editing Meet Structured Visuals},
  author={Zhuo, Le and Han, Songhao and Pu, Yuandong and Qiu, Boxiang and Paul, Sayak and Liao, Yue and Liu, Yihao and Shao, Jie and Chen, Xi and Liu, Si and others},
  journal={arXiv preprint arXiv:2510.05091},
  year={2025}
}

@article{ye2025unicedit,
  title={UnicEdit-10M: A Dataset and Benchmark Breaking the Scale-Quality Barrier via Unified Verification for Reasoning-Enriched Edits},
  author={Ye, Keming and Huang, Zhipeng and Fu, Canmiao and Liu, Qingyang and Cai, Jiani and Lv, Zheqi and Li, Chen and Lyu, Jing and Zhao, Zhou and Zhang, Shengyu},
  journal={arXiv preprint arXiv:2512.02790},
  year={2025}
}

@article{chow2025editmgt,
  title={EditMGT: Unleashing Potentials of Masked Generative Transformers in Image Editing},
  author={Chow, Wei and Li, Linfeng and Kong, Lingdong and Li, Zefeng and Xu, Qi and Song, Hang and Ye, Tian and Wang, Xian and Bai, Jinbin and Xu, Shilin and others},
  journal={arXiv preprint arXiv:2512.11715},
  year={2025}
}

@article{qian2025pico,
  title={Pico-banana-400k: A large-scale dataset for text-guided image editing},
  author={Qian, Yusu and Bocek-Rivele, Eli and Song, Liangchen and Tong, Jialing and Yang, Yinfei and Lu, Jiasen and Hu, Wenze and Gan, Zhe},
  journal={arXiv preprint arXiv:2510.19808},
  year={2025}
}

@article{ye2025echo,
  title={Echo-4o: Harnessing the power of gpt-4o synthetic images for improved image generation},
  author={Ye, Junyan and Jiang, Dongzhi and Wang, Zihao and Zhu, Leqi and Hu, Zhenghao and Huang, Zilong and He, Jun and Yan, Zhiyuan and Yu, Jinghua and Li, Hongsheng and others},
  journal={arXiv preprint arXiv:2508.09987},
  year={2025}
}

@article{sun2025unified,
  title={Unified continuous generative models},
  author={Sun, Peng and Jiang, Yi and Lin, Tao},
  journal={arXiv preprint arXiv:2505.07447},
  year={2025}
}

@article{lu2024simplifying,
  title={Simplifying, stabilizing and scaling continuous-time consistency models},
  author={Lu, Cheng and Song, Yang},
  journal={arXiv preprint arXiv:2410.11081},
  year={2024}
}

@article{lipman2022flow,
  title={Flow matching for generative modeling},
  author={Lipman, Yaron and Chen, Ricky TQ and Ben-Hamu, Heli and Nickel, Maximilian and Le, Matt},
  journal={arXiv preprint arXiv:2210.02747},
  year={2022}
}

@article{liu2026lumina,
  title={Lumina-mGPT: Flexible Photorealistic Autoregressive Text-to-Image Generation},
  author={Liu, Dongyang and Xin, Yi and Zhao, Shitian and Zhuo, Le and Lin, Weifeng and Li, Xinyue and Qin, Qi and Zhai, Guangtao and Liu, Xiaohong and Li, Hongsheng and others},
  journal=ijcv,
  year={2026}
}

@article{xin2025resurrect,
  title={Resurrect mask autoregressive modeling for efficient and scalable image generation},
  author={Xin, Yi and Zhuo, Le and Qin, Qi and Luo, Siqi and Cao, Yuewen and Fu, Bin and He, Yangfan and Li, Hongsheng and Zhai, Guangtao and Liu, Xiaohong and others},
  journal={arXiv preprint arXiv:2507.13032},
  year={2025}
}

@article{chen2024interleaved,
  title={Interleaved scene graphs for interleaved text-and-image generation assessment},
  author={Chen, Dongping and Chen, Ruoxi and Pu, Shu and Liu, Zhaoyi and Wu, Yanru and Chen, Caixi and Liu, Benlin and Huang, Yue and Wan, Yao and Zhou, Pan and others},
  journal={arXiv preprint arXiv:2411.17188},
  year={2024}
}

@inproceedings{zhou2025opening,
  title={Opening: A comprehensive benchmark for judging open-ended interleaved image-text generation},
  author={Zhou, Pengfei and Peng, Xiaopeng and Song, Jiajun and Li, Chuanhao and others},
  booktitle=cvpr,
  year={2025}
}

@article{liao2025mogao,
  title={Mogao: An omni foundation model for interleaved multi-modal generation},
  author={Liao, Chao and Liu, Liyang and Wang, Xun and Luo, Zhengxiong and Zhang, Xinyu and Zhao, Wenliang and Wu, Jie and Li, Liang and Tian, Zhi and Huang, Weilin},
  journal={arXiv preprint arXiv:2505.05472},
  year={2025}
}

@article{cui2025emu35nativemultimodalmodels,
      title={Emu3.5: Native Multimodal Models are World Learners}, 
      author={Yufeng Cui and Honghao Chen and Haoge Deng and Xu Huang and Xinghang Li and Jirong Liu and Yang Liu and Zhuoyan Luo and others},
      journal={arXiv preprint arXiv:2510.26583},
      year={2025}
}

\end{document}